\pdfoutput=1
\documentclass[11pt]{article}

\usepackage[final]{acl}
\usepackage{times}
\usepackage{latexsym}
\usepackage{mathtools}
\usepackage{float}
\usepackage{lipsum}  
\usepackage{amsmath}
\usepackage{amsthm}
\usepackage[pro]{fontawesome5}
\usepackage{amssymb}
\usepackage{makecell}
\usepackage{caption}
\usepackage{subcaption}

\newtheorem{theorem}{Theorem}
\newtheorem{corollary}{Corollary}[theorem]
\theoremstyle{definition}
\newtheorem*{remark}{Remark}

\DeclareRobustCommand{\logo}{%
  \begingroup\normalfont
  \raisebox{-0.3em}{%
  \hspace{-0.5em}
  \includegraphics[height=1.2em]{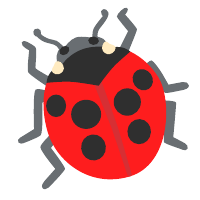}%
  }%
  \kern 0.2em
  \endgroup
}

\usepackage[T1]{fontenc}
\usepackage[utf8]{inputenc}

\usepackage{microtype}

\usepackage{inconsolata}

\usepackage{url}            % simple URL typesetting
\usepackage{hyperref}       % hyperlinks
\usepackage{booktabs}       % professional-quality tables
\usepackage{amsfonts}       % blackboard math symbols
\usepackage{nicefrac}       % compact symbols for 1/2, etc.
\usepackage{microtype}      % microtypography
\usepackage{xcolor}         % colors
\usepackage{enumitem}
\usepackage{graphicx}
\usepackage{enumitem}
\usepackage{multirow}
\usepackage{multicol}
\usepackage{tcolorbox}

\usepackage{titletoc}

\newcommand\modelshortShift{\text{{LoRMA$_{\pi}$}}}
\newcommand\modelshortRidge{\text{{LoRMA$_{+}$}}}

\title{\logo LoRMA: Low-Rank Multiplicative Adaptation for LLMs}

\author{
Harsh Bihany\thanks{Equal Contribution}
\qquad
Shubham Patel\footnotemark[1]
\qquad
Ashutosh Modi
\\ 
        Indian Institute of Technology Kanpur (IIT Kanpur)
\\
  \texttt{\{harshbi, devang, ashutoshm\}@cse.iitk.ac.in},  
}

\begin{document}
\maketitle
% Equal Contributions
% \def\thefootnote{*}\footnotetext{Equal Contribution.}\def\thefootnote{\arabic{footnote}}

% \begin{abstract}
% This document is a supplement to the general instructions for *ACL authors. It contains instructions for using the \LaTeX{} style files for ACL conferences. 
% The document itself conforms to its own specifications, and is therefore an example of what your manuscript should look like.
% These instructions should be used both for papers submitted for review and for final versions of accepted papers.
% \end{abstract}
\begin{abstract}
% \SP{
% \begin{itemize}
%     \item Avoid re-use text directly
% \end{itemize}
% }

% \lipsum[1]

Large Language Models have shown remarkable capabilities in the NLP domain. Their effectiveness can mainly be attributed to their ability to adapt to an array of downstream tasks. However, generally, full fine-tuning is a computationally expensive job. To mitigate this, many techniques have been developed that prime efficiency, a prominent one being Low-Rank Adaptation (LoRA). However, LoRA and its variants employ re-parametrized additive updates. In this paper, we propose Low-Rank Multiplicative Adaptation (LoRMA), which shifts the paradigm of additive updates to a richer space of matrix multiplicative transformations. We tackle challenges such as computational complexity and rank bottleneck of matrix multiplication by effectively re-ordering operations and introducing rank inflation strategies. We conduct extensive experiments to demonstrate the effectiveness of our approach in terms of various evaluation metrics.

% \SP{Shrink related work/background, use motivations and metrics to sell method}

% Their effectiveness can be attributed to their capability to be adapted to various downstream tasks through fine-tuning. 

% To mitigate this many techniques have emerged taking efficiency into account. 
% Techniques like LoRA have become pretty commonplace; 

\end{abstract}

% Plan
% \AM{
% Key contributions:
% \begin{itemize}
% \item Multiplicative method instead of Additive method (backed by proof, pre or post multiplication). But why multiplicative? 
% \item Motivation for Rank Inflation
% \item Two methods: Shift and Ridge Rank Inflation (backed by proof, complexity analysis, why random shift not possible?)
% \item Baseline Models: Adpators, BitFit, LoRA, AutoLora, DoRA
% \item Tasks: (RTE, SST2, STSB, MRPC, QNLI, MNLI, CoLA) AND (E2E, WebNLG)
% \item Roberta-Base, Roberta-Large, GPT-2 Medium
% \item Experiments: Baseline, Our methods, Ablation - the initial ranks of the weight matrices in pre-trained and fine-tuned models, presence / absence of rank inflation (pre/post, combination with LoRA, $W_{q}$ vs $W_{k}$, etc. ); rank progression, sub-space analysis, compare the final inflated matrices with LoRA matrices
% \item Results and Analysis 
% \end{itemize}
% }
% \SP{TODO: 
% \begin{itemize}
%     \item Dont forget to remove equal contribution footmark in anonymous version
%     \item Fix citations from arxiv to conference
% \end{itemize}
% }
\section{Introduction} 
\label{sec:intro}
Large Language Models (LLMs) have demonstrated strong performance across various NLP benchmarks \cite{open-llm-leaderboard-v2}. Though LLMs have shown impressive generalization capabilities (for example, via In-context learning \cite{dong2022survey}), sometimes these tend to have lower performance on some niche or low-resource tasks, thus requiring task-specific fine-tuning. LLMs usage follows a pre-train and fine-tune paradigm \cite{zhao2023survey}, where the model is trained on a massive amount of text in an unsupervised fashion, and subsequently, the model is fine-tuned for some specific tasks/domains in a supervised setting. Given the size of these models (order of billions of parameters), it may not always be feasible to fine-tune the entire model due to high computational costs. In recent years, a new class of techniques (referred to as PEFT (Parameter Efficient Fine Tuning)) has been proposed to address large computational costs associated with fine-tuning. 

\begin{figure}[t]
    \centering
    \includegraphics[width=0.80\linewidth]{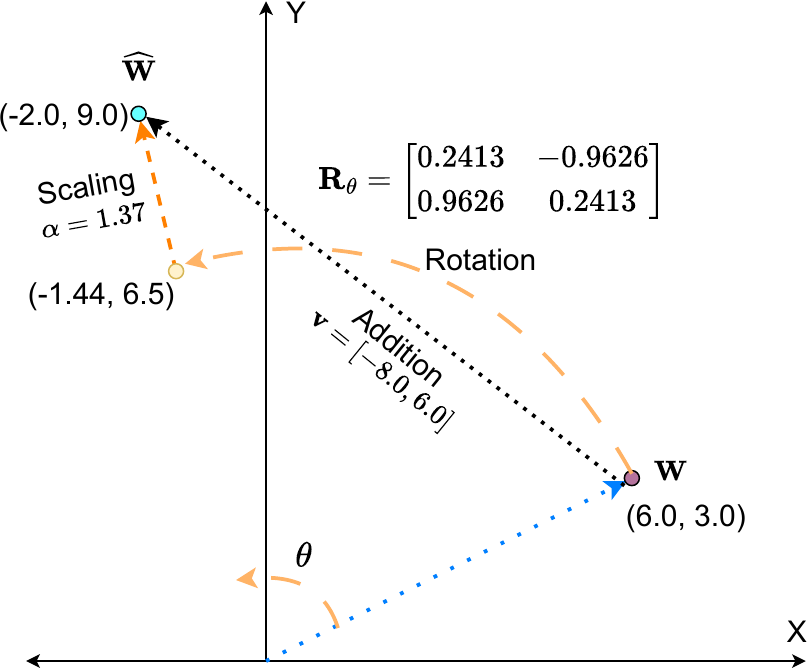}
    \caption{Transformation of a vector \textbf{W} by two methods: one is via rotation and scaling, the other is via the addition of a vector \textbf{v}.}
    \label{fig:add-vs-mult}
\end{figure}

\begin{figure*}[t]
    \centering
    \includegraphics[width=0.90\textwidth]{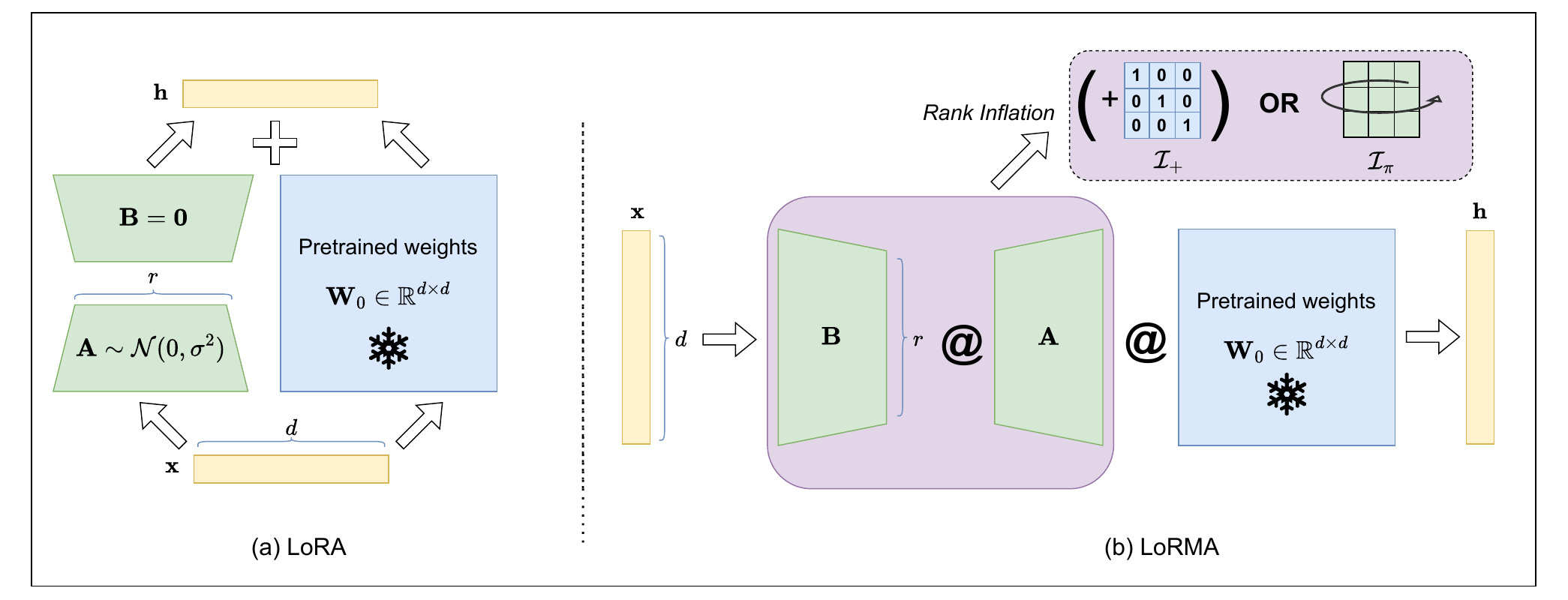}
    \caption{Comparing LoRA (a) and LoRMA\ (b). $@$ denotes matrix multiplication. $\mathcal{I}_{+}$ and $\mathcal{I}_{\pi}$ represent additive and permutation based rank inflation respectively (\S\ref{sec:methods}). In case of LoRMA, initialization of $\mathbf{A}$ and $\mathbf{B}$ depends on the type of inflation (\S\ref{sec:methods}).} 
    \label{fig:lorma}
\end{figure*}

\noindent Various PEFT techniques previously have been devised  \cite{han2024parameterefficient}; however, they often introduce trade-offs such as lack of parallelism, increased inference latency (e.g., Adaptors \cite{houlsby2019parameterefficienttransferlearningnlp}), or restricted sequence lengths \cite{petrov2024promptingprefixtuningworktheory}. Consequently, re-parametrization-based techniques such as Low-Rank Adaptation (LoRA) \cite{hu2022lora} based fine-tuning methods have gained popularity. Typically, during fine-tuning, the weights (in the form of the weight matrix, e.g., query/key/value matrix) of LLMs are updated using additive update rule, i.e., $\mathbf W_0 + \Delta \mathbf W$, where $\Delta \mathbf W$ is the update in the weights obtained due to fine-tuning. The main idea behind LoRA is to approximate the update matrix $\Delta \mathbf W \in \mathbb{R}^{d \times k}$ by a low-rank approximation $\frac{\alpha}{r}\cdot\mathbf{B} \mathbf{A}$, where $\mathbf{B} \in \mathbb{R}^{d \times r}$ and $\mathbf{A} \in \mathbb{R}^{r \times k}$ are low-rank matrices ($r \ll d, k$), $\frac{\alpha}{r}$ is a scaling factor, leading to $\mathbf W = \mathbf W_0 + \frac{\alpha}{r} \cdot \mathbf{B} \mathbf{A} $. It is based on the study that the additional information required for task-specific updates has a smaller intrinsic rank and lies on a much smaller manifold compared to the entire space of $d \times k$ matrices \cite{aghajanyan-etal-2021-intrinsic,hu2022lora}. The current LoRA-based approaches \cite{yang2024low} have employed additive transformations, where the low-rank update matrix can be added to the original weight matrix during inference. However, a similar transformation could also be achieved via multiplicative updates. For example, consider a weight vector $\mathbf W$ (Fig. \ref{fig:add-vs-mult}) and we would like to transform it to vector $\widehat{\mathbf W}$, this could be accomplished via the addition of a vector $\mathbf{v}$, or it could also be done by rotating $\mathbf W$ by angle $\theta$ (done via Rotation Matrix $\mathbf{R_{\theta}}$) and subsequently by scaling it by scalar $\alpha$. 

\noindent Inspired by this, we propose Low-Rank Multiplicative Adaptation (LoRMA) for efficiently fine-tuning LLMs on new tasks. LoRMA applies low-rank multiplicative update to a weight matrix, i.e., $\mathbf W = \frac{\alpha}{r} \cdot (\mathbf{B}\mathbf{A}) \mathbf W_0$, where $\frac{\alpha}{r}$ is a scalar and $\mathbf{A} \in \mathbb{R}^{d \times r}$ and $\mathbf{B} \in \mathbb{R}^{r \times d}$ are low-rank matrices ($r \ll d, k$). However, this simple multiplicative update faces two new challenges: an increase in computational complexity due to additional matrix multiplication operations and a restriction on the maximum rank of $\mathbf{W}$ due to the property: $\mathcal R(\mathbf{AB}) \leq \min (\mathcal R(\mathbf{A}), \mathcal R(\mathbf{B}))$, where $\mathcal R(\cdot)$ denotes the rank of a matrix. We employ appropriate ordering of matrix multiplication to address the issue of computational complexity (\S\ref{sec:methods}). Additionally, to counteract the issue of \textit{rank inhibition} caused by matrix multiplication, we introduce \textit{rank inflation} strategies and demonstrate their effectiveness (Fig. \ref{fig:lorma}). On average, the proposed techniques have better performance than LoRA (\S\ref{sec:experiments}). Moreover, it has a much faster convergence rate (hence lower training time) as compared to LoRA (\S\ref{sec:results}). 

\noindent In a nutshell, we make the following contributions:
\begin{itemize}[noitemsep,nosep,leftmargin=*]
\item We propose a new PEFT technique for adapting LLMs for downstream tasks: Low-Rank Multiplicative Adaptation (LoRMA). We employ multiplicative updates as an alternative to additive updates used in LoRA. To make the proposed method computationally efficient and overcome rank inhibition brought in by matrix multiplication of low-rank matrices, we propose two variants: Low-Rank Multiplicative Adaptation with additive inflation (\modelshortRidge) and Low-Rank Multiplicative Adaptation with permutation-based inflation (\modelshortShift). We propose a generic framework that can adapted into existing variants of LoRA such as Q-LoRA \cite{dettmers2023qlora}, AutoLoRA \cite{zhang-etal-2024-autolora}, DyLoRA \cite{valipour-etal-2023-dylora}, and DoRA \cite{liu2024dora}).
\item We perform an extensive set of experiments on transformer-based LLMs (RoBERTa, GPT-2, Gemma-2B, and LLaMA-3-8B) on various NLU and NLG tasks and compare them with existing baselines. On average, the proposed techniques perform better. We show that LoRMA shows faster convergence. Via various ablation studies, we demonstrate the benefits of the approach and analyze the effect of rank, weight matrix choice, and correlation between weight updates of LoRA and LoRMA. We release our code at \url{https://exploration-lab.github.io/LoRMA/}. %\url{https://github.com/Exploration-Lab/LoRMA}.
\end{itemize}

\section{Related Work} \label{sec:related}

LLMs are generally fine-tuned using Parameter Efficient Fine Tuning (PEFT) methods. Existing PEFT techniques typically fall into three categories \cite{han2024parameterefficient}: (1) Additive methods (these involve the inclusion of a small set of additional trainable parameters/modules, e.g., Adaptors \cite{houlsby2019parameterefficienttransferlearningnlp}, Prefix-tuning \cite{li-liang-2021-prefix}); (2) Selective methods (these involve selecting a smaller subset of parameters/modules (e.g., bias in the case of BitFit \cite{zaken2021bitfit}) and fine-tuning only those (via application of binary masks), e.g., Diff pruning \cite{guo2020parameter}); (3) Re-parametrization techniques (these involve re-parameterization of existing weight update matrix via low-rank approximation, e.g., LoRA \cite{hu2022lora}). In this paper, we focus on re-parameterization-based approaches.

\noindent Several variants of LoRA have been proposed \cite{lora_plus, hydralora}, each focusing on different aspects of the method \cite{mao2025survey}. Here, we describe some of the prominent ones; for more details, please refer to the survey by \citet{yang2024low}. DyLoRA \cite{valipour-etal-2023-dylora} dynamically searches for optimal ranks for different weight matrices of the model rather than using a fixed rank across all layers. Methods like AutoLoRA \cite{zhang-etal-2024-autolora} and AdaLoRA \cite{zhang2023adaptive} adaptively allocate the parameter budget across the model matrices by determining an importance score. ReLoRA \cite{lialin2024relora} introduces aggregated low-rank updates to large neural networks during the training phase with a jagged learning rate scheduler, which depends on the interval in which updates are made to the weight matrix. DoRA \cite{liu2024dora} improves convergence by splitting magnitude and directional updates, enabling weight updates close to traditional fine-tuning. VeRA \cite{kopiczko2024vera} further reduces storage requirements by using fixed matrices $\mathbf{A}$ and $\mathbf{B}$ across layers and introducing trainable diagonal matrices. SVFT \cite{svft} performs a singular value decomposition of the weight matrix and modifies the singular-value matrix using a trainable diagonal (SVFT$^P$ - Plain SVFT) or by selecting trainable elements randomly (SVFT$^R_d$). PRoLoRA \cite{wang-etal-2024-prolora} introduces re-using parameters within the LoRA adapter matrix by replicating chunks across rows and columns. The paper introduces a \textit{rotation enhancement} operation involving chunks in the adapter matrices to recover the expressivity in $\mathbf{B}  \mathbf{A}$ lost due to replicating parameters and add a set of trainable parameters to further enhance expressivity. Our work is different from PRoLoRA; we introduce operations at the row level to inflate the rank of the matrix.
%While it appears similar to $\mathcal{I}_{\pi}$ operation (\S\ref{sec:methods}) that we independently propose in our methodology, $\mathcal{I}_{\pi}$ operates at the row level and is introduced with the motivation of inflating the rank of the matrix. Unlike PRoLoRa, due to our multiplicative paradigm, we also develop an initialization strategy to make $\mathcal{I}_{\pi}(\mathbf{B} \mathbf{A}) = \mathbf I$ during the start of the training procedure. 

\noindent Most of these methods discussed are additive in nature, with the exception of SVFT. We explore the effect of replacing additive modules with \textit{multiplicative transformations}. By investigating multiplicative updates, we aim to address some of the limitations of additive approaches while maintaining the efficiency and effectiveness of PEFT. Multiplicative updates offer a more expressive mechanism for modifying weight matrices. By leveraging matrix multiplication, we can encode richer transformations, which may better capture several complex relationships.
We propose a generic multiplicative variant of the additive LoRA. Our proposed variant is orthogonal to many of these variants, thus enabling one to further improve the strategy's effectiveness by combining our variant with existing variants in the literature. For example, analogous to efficient rank-allocation strategies like AutoLoRA for additive LoRA, an equivalent multiplicative variant like AutoLoRMA can be devised by transforming the weight update to be multiplicative and using the rank-allocation strategy of AutoLoRA. Similarly, approaches can be devised for QLoRMA, AdaLoRMA, etc. Given this motivation, we primarily benchmark our proposed approach of LoRMA against LoRA to show that it has a competitive performance.
\section{Methodology} 
\label{sec:methods}

\subsection{Background} \label{sec:background}

\noindent Rank of a matrix ($\mathcal{R}(\cdot)$) is defined as the number of linearly independent rows/columns of a matrix and is equivalent to the dimensionality of the space spanned by the rows/columns of the matrix. The rank of a matrix is a fundamental quantity that captures various important characteristics. Some of the key properties \cite{strang2022introduction} are: 

\begin{tcolorbox}[boxrule=0.5pt, top=0pt, bottom=0pt]
{\small
\parbox{\linewidth}{
\setlength{\abovedisplayskip}{3pt}
\setlength{\belowdisplayskip}{3pt}
\begin{align}  
&\mathcal{R}(\mathbf{M}) \ \le \  \min(n, m), \text{ for }  \mathbf{M} \in \mathbb{R}^{n \times m} \label{eqn:dimension} \\
 &\mathcal{R}(\mathbf M_1 + \mathbf M_2) \ \geq \  |\mathcal{R}(\mathbf M_1) - \mathcal{R}(\mathbf M_2)| \label{eqn:sum} \\
 &\mathcal{R}\left(\mathbf M_1 \times \mathbf M_2\right) \ \le \  \min (\mathcal{R}(\mathbf M_1), \mathcal{R}(\mathbf M_2)) \label{eqn:product} \\
 &\mathcal{R}(\mathbf{M}) \ = \ n,\  \mathbf{M} \in \mathbb{R}^{n \times n} \text{ if }\mathbf{M} \text{ is invertible} \label{eqn:inverse}
\end{align}
}%parbox
}%small
\end{tcolorbox}
% \setlength{\abovedisplayskip}{1pt}
% \setlength{\belowdisplayskip}{1pt}
% % Dimension
% \begin{equation} \label{eqn:dimension}
% \scriptstyle
%     \mathcal{R}(\mathbf{M}) \ \le \ \min(n, m),\  \mathbf{M} \in \mathbb{R}^{n \times m} 
% \end{equation}
% % Sum
% \begin{equation} \label{eqn:sum}
% \scriptstyle
%     \mathcal{R}(\mathbf{M_1} + \mathbf{M_2}) \ \geq \  |\mathcal{R}(\mathbf{M_1}) - \mathcal{R}(\mathbf{M_2})|
% \end{equation}
% % Product
% \begin{equation} \label{eqn:product}
% \scriptstyle
%     \mathcal{R}\left(\mathbf{M_1} \times \mathbf{M_2}\right) \ \le \  \min (\mathcal{R}(\mathbf{M_1}), \mathcal{R}(\mathbf{M_2}))
% \end{equation}
% %inverse
% \begin{equation} \label{eqn:inverse}
% \scriptstyle
% \mathcal{R}(\mathbf{M}) \ = \ n,\  \mathbf{M} \in \mathbb{R}^{n \times n} \text{ if }\mathbf{M} \text{ is invertible}
% \end{equation}
\noindent Property \ref{eqn:dimension} indicates that the rank of a matrix is bounded by its dimensions. Property \ref{eqn:sum} specifies a lower bound for the rank when matrices undergo addition. Property \ref{eqn:product} constrains the rank of the product of two matrices to be bounded by the smaller of both. Property \ref{eqn:inverse} states that square matrices that are invertible (for example, identity matrix $\mathbf{I}_n$) have a rank equal to the number of rows/columns ($n$).

\noindent LoRA \cite{hu2022lora} updates a pre-trained weight matrix $\mathbf W_0 \in \mathbb R^{d \times k}$ by additive update, i.e., $\mathbf h = \mathbf{W}_0\mathbf{x} + \Delta \mathbf{Wx}$, where $\mathbf{x}$ is the input and $\mathbf W_0$ is frozen during fine-tuning. The updates $\Delta \mathbf{W}$ are constrained to a low-rank decomposition $\mathbf B \mathbf A$ where $\mathbf B \in \mathbb R^{d \times r}, \mathbf A \in \mathbb R^{r \times k}$ and $r \ll \min(d, k)$, i.e.,  
\begin{align}
    \mathbf h = (\mathbf W_0 + \underbrace{\frac{\alpha}{r} \cdot \mathbf B\mathbf A)}_{\Delta \mathbf{W}} \mathbf x
\label{eqn:lora}
\end{align}
% {
% \parbox{\linewidth}{
% \setlength{\abovedisplayskip}{3pt}
% \setlength{\belowdisplayskip}{3pt}
% \begin{align*}  
% \mathbf h = (\mathbf W_0 + \underbrace{\frac{\alpha}{r} \cdot \mathbf B\mathbf A)}_{\Delta \mathbf{W}} \mathbf x  \label{eq:lora}
% \end{align*}
% }%parbox
% }%small

\noindent where $\alpha$ is a scalar. To ensure that the initial training pass resembles the pre-trained model and training stability, $\mathbf B$ is initialized to $\mathbf 0$. 

%\subsection{Existence} 
\noindent\textbf{Existence:}
In LoRA, weights are updated via additive updates; however, we are proposing a different paradigm where weights are updated via a multiplicative process. One could argue if it is even feasible to attain the same updates via a multiplicative process. In this regard, we first provide proof that it is indeed possible to transform a matrix into another matrix via multiplicative mapping. 

% \begin{tcolorbox}[boxrule=0.5pt, top=0pt, bottom=0pt]
% %{\small
% \begin{theorem}
% Given matrices $\mathbf{M_0}, \mathbf{M} \in \mathbb{R}^{n \times m}$, where $n > m$ and $\mathcal{R}(\mathbf{M_0})=m$, there exists a matrix $\mathbf{M_A} \in \mathbb{R}^{n\times n}$ having adaptable parameters that can transform $\mathbf{M_0}$ into $\mathbf{M}$ via pre-multiplication, i.e., $\mathbf{M} = \mathbf{M_A M_0}$.  However, post-multiplication with $\mathbf{M_A} \in \mathbb{R}^{m\times m}$ does not guarantee the existence of $\mathbf{M}$, i.e., $P(\mathbf{M} \neq \mathbf{M_0 M_A}) > 0 \ \forall \ \mathbf{M_A} \in \mathbb{R}^{m\times m}$, where $P(.)$ is the probability measure.  
% \end{theorem}
% %}%small
% \end{tcolorbox}
% \begin{tcolorbox}[boxrule=0.5pt, top=0pt, bottom=0pt]
% %{\small
% \begin{theorem}
% Given matrices $\mathbf{M_0}, \mathbf{M} \in \mathbb{R}^{n \times m}$, where $n > m$ and $\mathcal{R}(\mathbf{M_0})=m$, there always exists a matrix $\mathbf{M_A} \in \mathbb{R}^{n\times n}$ that can transform $\mathbf{M_0}$ into $\mathbf{M}$ via pre-multiplication, i.e., $\mathbf{M} = \mathbf{M_A M_0}$.  However, post-multiplication with $\mathbf{M_A} \in \mathbb{R}^{m\times m}$ does not guarantee the existence of $\mathbf{M}$, i.e., $\exists M_0, M \in \mathbb{R}^{n \times m}\
% such\ that\ \forall M_A \in \mathbb{R}^{n \times n}, M \ne M_A M_0$ 
% \end{theorem}
% %}%small
% \end{tcolorbox}

%{\small
% \begin{tcolorbox}[boxrule=0.5pt, top=0pt, bottom=0pt]
\begin{theorem} \label{thm:main_thm}
Given $\mathbf M_0 \in \mathbb R^{n \times m}$ where $n > m$ and let $\mathcal R(\mathbf M_0) = m$. For all $\mathbf M \in \mathbb R^{n \times m}, \exists\ \mathbf M_A \in \mathbb R^{n \times n},$ such that $\mathbf M = \mathbf M_A \mathbf M_0$.
\end{theorem}
% \end{tcolorbox}
%}%small

\begin{proof}
Given $\mathbf M_0 \in \mathbb R^{n \times m}$ where $n > m$ and $\mathcal R(\mathbf M_0) = m$, implies that $\mathbf M_0$ is a full column matrix, i.e., all its columns are independent. This implies that there exists a left inverse of the matrix $\mathbf M_0$, say $\mathbf M_0^+$, such that $\mathbf M_0^+ \mathbf M_0 = \mathbf I_m$. We need to show the existence of a matrix $\mathbf M_A$ for any given $\mathbf M \in \mathbb R^{n \times m}$, such that \textit{pre-multiplication} of $\mathbf M_A$ with $\mathbf M_0$ gives $\mathbf M$, i.e., $\mathbf M = \mathbf M_A \mathbf M_0$. Construct the matrix $\mathbf M_A = \mathbf M \mathbf M_0^+$. This proves the claim as $\mathbf M_A \mathbf M_0 = (\mathbf M \mathbf M_0^+)  \mathbf M_0 = \mathbf M \mathbf I_m = \mathbf M$.
\end{proof}

\begin{corollary}
    Given $\mathbf M_0 \in \mathbb R^{n \times m}$ where $n > m$ and $\mathcal R(\mathbf M_0) = m$. There exists $\mathbf M \in \mathbb R^{n \times m}$ such that $\forall\ \mathbf M_A \in \mathbb R^{m \times m}, \mathbf M \neq \mathbf M_0 \mathbf M_A$.
\end{corollary}

\begin{proof}
    Suppose that $\forall\ \mathbf M \in \mathbb{R}^{n \times m}, \exists\ \mathbf M_A$, such that \textit{post-multiplication}, i.e., $\mathbf M_0 \mathbf M_A = \mathbf M$. In other words $\mathbb R^{n \times m} = \{ \mathbf M_0\mathbf M_A\,|\, \mathbf M_A \in \mathbb R^{m \times m} \}$. This does not hold as the \textit{degrees of freedom} on the right-hand side for a given full column matrix $\mathbf M_0$ is $m^2$ (number of elements in $\mathbf M_A$), while the potential \textit{degrees of freedom} required is $nm$-many in $\mathbb R^{n \times m}$. Formally, consider a counter-example. Assume the given $\mathbf M_0 = \begin{pmatrix}
        \mathbf I_m \\ \mathbf 0_{n-m}
    \end{pmatrix}$. Let the required transformation be to $\mathbf M = \begin{pmatrix}
        \mathbf 0_{n-m} \\ \mathbf I_m
    \end{pmatrix}$, where $\mathbf 0_{n-m}$ denotes a zero matrix $\in \mathbb R^{(n-m) \times m}$. It is easy to verify that $\nexists\ \mathbf M_A \in \mathbb R^{m \times m}$ which satisfies the desired transformation.
\end{proof}

% 
% 
% Proof 2 Begin
% 
% 
% \begin{figure}[htb]
%     \centering
%     \includegraphics[width=\linewidth]{images/small-proof.pdf}
%     \caption{Counterproof for smaller dimension}
%     \label{fig:small-proof}
% \end{figure}

% \noindent \underline{1:} The adapter matrix $\mathbf{M_A}$ is being pre-multiplied to $\mathbf{M_0}$ on the side of the smaller dimension $m$. $\mathbf{M} = \mathbf{M_0} \times \mathbf{A}$.

% \noindent \underline{2:} Each column $\mathbf{M}$ is a weighted sum of the columns of $\mathbf{M_0}$ as follows: For $j \in [1,m]:$
% \[
% \mathbf{M}[:, j] = \sum\limits_{k=1}^{m} \alpha_{kj}\mathbf{M_0}[:, k] \text{ with } \alpha_{kj} = \mathbf{M_A}[k,j] 
% \]
% \noindent \underline{3:} If any column $\mathbf{M}[j,:] \in \mathbb{R}^{n}, n>m$ does not belong to the $m$ dimensional space of the linear span of the columns of $\mathbf{M}$, it can not be expressed in a form as above. Hence for such $\mathbf{M}$, $\forall\ \mathbf{M_A} \in \mathbb{R}^{n \times m},\ \mathbf{M} \ne \mathbf{M_A M}$ 
% 
% 
% Proof 2 End
% 
% 

\begin{corollary} \label{cor:square_mult}
    Given the square matrix $\mathbf{M} \in \mathbb{R}^{n \times n}$ and non-singular matrix $\mathbf{M}_0 \in \mathbb{R}^{n \times n}$, there exist matrices $\mathbf{M}_{A_\ell}, \mathbf{M}_{A_r} \in \mathbb{R}^{n \times n}$ that can transform $\mathbf M_0$ into $\mathbf{M}$ via \textit{pre-multiplication/post-multiplication} respectively, i.e., $\mathbf{M} = \mathbf M_{A_\ell} \mathbf M_0$ and $\mathbf{M} = \mathbf M \mathbf M_{A_r}$. 
\end{corollary}

\begin{remark}

We present the above results to motivate the existence of a multiplicative transformation that maps frozen pre-trained weight matrices $\mathbf{M}_0$ to potentially any other set of weights with the same dimensionality. 
A key requirement underlying this hypothesis is that the weight matrices---such as $\texttt{attention.self.query}$ in RoBERTa or the spliced $\texttt{c\_attn}$ in GPT-2 (the models used in \S\ref{sec:experiments})---are invertible. 
To ensure this, we verify that these matrices are either full rank or close to full rank, typically within $99\%$ of the maximum possible rank.

\end{remark}

\subsection{LoRMA}
\noindent Theorem \ref{thm:main_thm} guarantees the existence of a matrix $\mathbf M_A$ for a desired transformation. Hence, we propose a multiplicative update rule, i.e., $\mathbf M_A \times \mathbf{W}_0 $. The update is approximated using low-rank approximation, i.e., 
\begin{equation}
    \mathbf h = ((\mathbf{B} \mathbf{A}) \times \mathbf W_0) \mathbf x 
\label{eqn:lorma}
\end{equation}
% 
% {
% \parbox{\linewidth}{
% \setlength{\abovedisplayskip}{3pt}
% \setlength{\belowdisplayskip}{3pt}
% \begin{align*}  
% \mathbf h = ((\mathbf{B} \mathbf{A}) \times \mathbf W_0) \mathbf x 
% \end{align*}
% }%parbox
% }%small

\noindent where, $\mathbf{B} \in \mathbb{R}^{d \times r}$, $\mathbf{A} \in \mathbb{R}^{r \times d}$ with $r \ll \min(d,k)$ are low-rank matrices such that the product $\mathbf{BA}$ captures the desired transformation of matrix $\mathbf W_0$. However, this naive approach has a few shortcomings. In accordance with property \ref{eqn:product}, the resultant matrix product is limited to be of rank $r$ since $\mathcal{R}(\mathbf{BAW}_0) \le \mathcal{R}(\mathbf{B}) \leq r$. This significantly undermines the potential desirable independence of rows/columns in the final representation of the updated weights. Further, during the onset of the fine-tuning, in the case of LoRA, it is preferable to have $\Delta \mathbf{W} = \mathbf{0}$, so that $\mathbf h = \mathbf{W} \mathbf{x}$, this ensures stability during fine-tuning \cite{hu2022lora}. This is achieved by initializing $\mathbf{B}$ with zeros, ensuring that the additive update starts at zero. In our case, this would require the matrix $\mathbf{BA}$ to be equal to the identity matrix $\mathbf{I}_d$. However, the property \ref{eqn:product} dictates that this cannot be the case as $\mathcal R(\mathbf I_d) = d$. This forces the tuning to have a significant deviation from the beginning. We propose two strategies to mitigate the rank limitation imposed by low-rank matrices to capture the multiplicative transformation. 

\begin{figure}[t]
    \centering
    \includegraphics[scale=0.25]{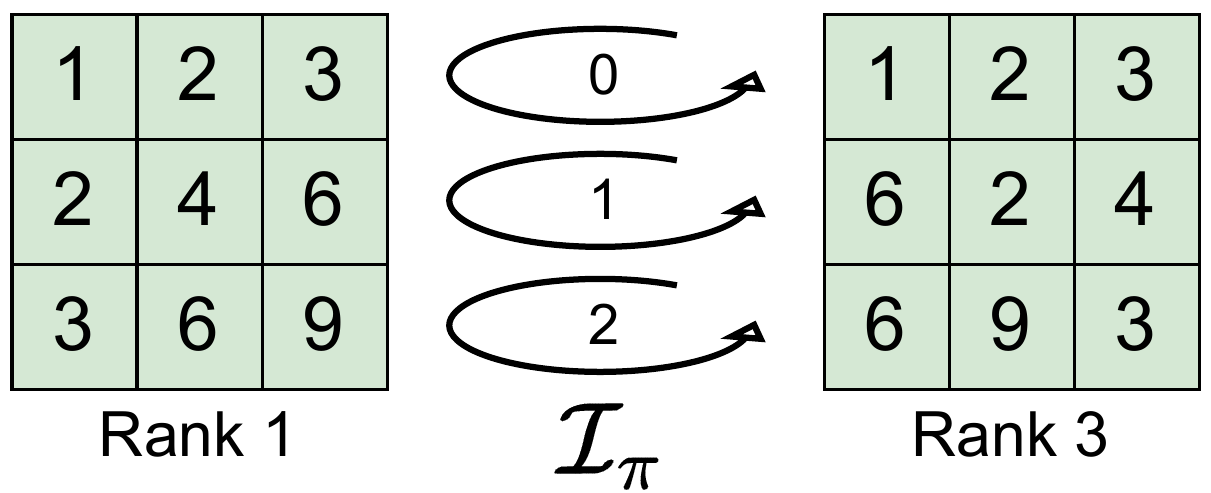}
    \vspace{-3mm}
    \caption{Permutation-Based Inflation $\mathcal{I}_{\pi}$ operation. Re-arrange matrix entries to inflate the rank.}
    \label{fig:shift}
    \vspace{-5mm}
\end{figure}

\subsubsection{Permutation-Based Inflation ($\mathcal{I}_{\pi}$)} 
Permutation-based rank inflation utilizes the idea of strategic re-arrangement of elements of the matrices to increase the rank of a matrix. The rows of the matrix are rotated cyclically in incremental steps. The $i$ th row is rotated by $i$, i.e. (row 0 by 0, row 1 by 1 ...). As can be seen in Fig. \ref{fig:shift}, this effective rearranging of a matrix's elements has enhanced the matrix's rank from 1 to a full rank of 3. We introduce this operation on the product of the matrices $\mathbf{BA}$, which equips the model with the ability to learn a higher-rank representation. Since the operation is simply a re-arrangement of the parameters, it does not make the gradient in-tractable.
% to increase its rank from $r$ and hence inflate the rank of the resultant weight $\mathcal{I}_{\pi}(\mathbf{BA})\mathbf{W_0}$. 
\begin{equation}
    \mathbf h = (\mathcal{I}_{\pi}(\mathbf{BA}) \times \mathbf W_0) \mathbf x
\label{eqn:lorma_pi}
\end{equation}
% {
% \parbox{\linewidth}{
% \setlength{\abovedisplayskip}{3pt}
% \setlength{\belowdisplayskip}{3pt}
% \begin{align*}  
% \mathbf h = (\mathcal{I}_{\pi}(\mathbf{BA}) \times \mathbf W_0) \mathbf x 
% \end{align*}
% }%parbox
% }%small

\noindent This inflation strategy also provides a better initialization scheme. This is achieved by warranting $\mathcal{I}_{\pi}(\mathbf{BA}) = \mathbf{I}_d$. The first column of $\mathbf{B}$ is set to ones, while the rest of the elements are randomly initialized. $\mathbf{A}[0, 0]$ is set to one, while the rest of the elements in $\mathbf{A}$ are set to zero. We refer to this variant as \modelshortShift. 

\subsubsection{Additive Rank Inflation ($\mathcal{I}_{+}$)}
Motivated by the need for an identity initialization of the transformation matrix, we introduce another technique to address the rank limitation inherent in low-rank approximations. Drawing inspiration from ridge regression, where the solution is stabilized by adding a regularization term $\left(\hat{\mathbf{\theta}} = (\mathbf{X}^T \mathbf{X} + \lambda \cdot \mathbf{I})^{-1}\mathbf{X^T} \mathbf{Y}\right)$, we incorporate an identity matrix into our formulation through addition. Specifically, the resulting transformation takes the form:
\begin{equation}
    \mathbf h = \mathcal{I}_{+}(\mathbf{BA})\mathbf{W}_0 \mathbf x = \left( \frac{\alpha}{r} \cdot \mathbf{BA} + \mathbf{I}_d\right) \mathbf{W}_0 \mathbf x
\label{eqn:lorma_plus}
\end{equation}
% 
% % 
% {
% \setlength{\abovedisplayskip}{3pt}
% \setlength{\belowdisplayskip}{3pt}
% \begin{align*}  
% \mathbf h = \mathcal{I}_{+}(\mathbf{BA})\mathbf{W}_0 \mathbf x = \left( \frac{\alpha}{r} \cdot \mathbf{BA} + \mathbf{I}_d\right) \mathbf{W}_0 \mathbf x
% \end{align*}
% }%small
% % 
% 
\noindent The rank of the sum $\left(\frac{\alpha}{r} \cdot \mathbf{BA} + \mathbf{I}_d\right)$ (here $\alpha$ is the scaling factor) is guaranteed to be at least $d - r$, as dictated by property \ref{eqn:sum}. Since $r \ll d,\ d-r \approx d$, this preserves sufficient rank flexibility, enabling richer transformations during training. This approach ensures that the transformation begins with identity initialization at the start of the fine-tuning process by setting $\mathbf{B}=\mathbf{0}$ and randomly initializing $\mathbf A$. We refer to this variant as \modelshortRidge.

\noindent To summarize, formally, the update rule for LoRMA is given by:

\begin{equation}
    \mathbf h = (\mathcal{I}(\mathbf{B} \mathbf{A}) \times \mathbf{W}_0) \mathbf x
\label{eqn:lorma_inflated}
\end{equation}

% {
% \parbox{\linewidth}{
% \setlength{\abovedisplayskip}{3pt}
% \setlength{\belowdisplayskip}{3pt}
% \begin{align*}  
% \mathbf h = (\mathcal{I}(\mathbf{B} \mathbf{A}) \times \mathbf{W}_0) \mathbf x 
% \end{align*}
% }%parbox
% }%small

\noindent where, $\mathbf W_0 \in \mathbb R^{d \times k}$, $\mathbf{B} \in \mathbb{R}^{d \times r}$, $\mathbf{A} \in \mathbb{R}^{r \times d}$ and $r \ll \min(d, k)$ and $\mathcal{I}$ denotes rank inflation techniques employed $\left( \mathcal{I}_{\pi} / \mathcal{I}_{+} \right)$. $\mathbf{A}$ and $\mathbf{B}$ are initialized such that $\mathcal{I}(\mathbf{BA}) = \mathbf{I}_d$. In our case, the application of the LoRMA over RoBERTa, GPT-2, Gemma-2B, and LLaMA-3-8B (\S\ref{sec:experiments}) is over square matrices and Corollary~\ref{cor:square_mult} ensures the existence of a multiplicand which is being adapted.

\begin{table}[t]
    \small
    \centering
	\begin{tabular}{lcc}
		\toprule
		Method & Computation & Complexity \\
		\midrule
            LoRA & $(\mathbf W_0 + \mathbf{BA})\mathbf{x}$ & $\mathcal{O}(dkb)$ \\
            LoRMA & $\mathbf{BAW}_0\mathbf{x}$ & $\mathcal{O}(dkb)$ \\
            \modelshortShift & $\mathcal{I}_{\pi}(\mathbf{BA}) \mathbf W_0\mathbf{x}$ & $\mathcal{O}(d^2(r + b))$ \\
            \modelshortRidge & $\mathbf W_0\mathbf{x} + \mathbf{BA} \mathbf W_0\mathbf{x}$ & $\mathcal{O}(dkb)$ \\
		\bottomrule
	\end{tabular}
    \vspace{-3mm}
    \caption{Time Complexity for computations incurred by different methods during training time.}
	\label{tab:time-complexity}
    \vspace{-5mm}
\end{table}

\subsubsection{Time Complexity and Advantages}
An obvious consideration to take is the computational cost incurred by the multiplicative transformations that are being introduced. Table~\ref{tab:time-complexity} (also see App. \S\ref{app:time-complexity}) provides a comparative analysis of the computational costs of LoRA for $\mathbf{x} \in \mathbb{R}^{k \times b}$ where $b$ denotes the batch size. Utilizing associativity of matrix multiplications and first performing multiplication with $\mathbf{x}$ helps make the cost of LoRMA comparable to LoRA. The cost of \modelshortShift\ is slightly higher since there is the requirement to first compute $\mathbf{BA}$ since the $\mathcal{I}_{\pi}$ operation is being applied on the product. 
\begin{figure}[htbp]
    \centering
    \includegraphics[width=0.9\linewidth]{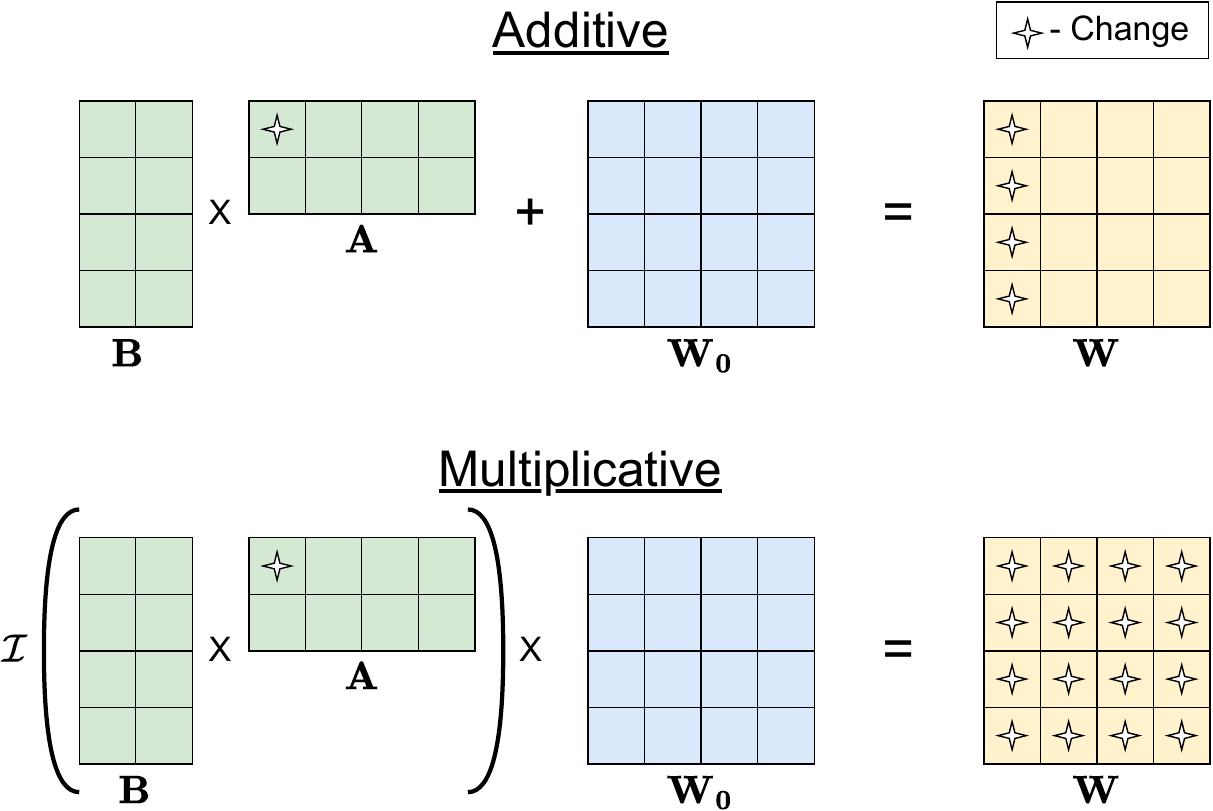}
    \vspace{-3mm}
    \caption{Impact on the resultant matrix on updating a single element in Additive vs Multiplicative updates.}
    \label{fig:large-impact}
    \vspace{-5mm}
\end{figure}
The advantages of LoRMA, similar to LoRA, include avoiding inference-time latency by permitting the merging of updates into the frozen weights, i.e., $\mathbf{W}_{\text{fine-tuned}} = \mathcal{I}(\mathbf{BA}) \times \mathbf W_0$. In the multiplicative representation, on updating a single parameter, the resultant weight matrix has many more updates as compared to additive transformations, as can be seen in Fig. \ref{fig:large-impact}. This can lead to the requirement of fewer updates to modify the weight matrix to another matrix, leading to faster convergence. We observe this empirically in our experiments (\S\ref{subsec:faster_conv}). Also, as compared to the restricted low-rank weight-updates for LoRA, LoRMA$_\pi$ has nearly full-rank (\S\ref{subsec:weight-update-comparision}) and hence richer update.

\section{Experimentation}  \label{sec:experiments}

\begin{table*}[htb]
  \centering
  % \footnotesize
  \small
  \addtolength{\tabcolsep}{-4pt}
  \begin{tabular}{l c ccccccccc}
  \hline
  \toprule
  Method & \# Params & MNLI & SST-2 & MRPC & CoLA & QNLI & QQP & RTE & STS-B & Avg. \\
  \midrule

    RoBERTa\textsubscript{base} (FT)* & 125.0M & 87.6 & 94.8 & 90.2 & 63.6 & 92.8 & 91.9 & 78.7 & 91.2 & 86.4 \\
    \midrule
    RoBERTa\textsubscript{base} (BitFit)* & 0.1M & 84.7 & 93.7 & \textbf{92.7} & 62.0 & 91.8 & 84.0 & \textbf{81.5} & \underline{90.8} & 85.2 \\
    % RoBERTa\textsubscript{base} (AutoLoRA)* & 0.3M & 87.0 & \textbf{94.9} & 89.4 & 61.3 & \textbf{92.9} & 90.3 & 77.0 & \underline{90.8} & 85.5 \\
    RoBERTa\textsubscript{base} (LoRA) & 0.3M & \textbf{87.5} & 94.6 & 91.0 & \underline{63.6} & \underline{92.7} & \textbf{90.8} & \underline{78.0} & 89.5 & \underline{85.9} \\
    RoBERTa\textsubscript{base} (\modelshortShift) & 0.3M & \underline{87.4} & 94.2 & 91.1 & 63.5 & 92.1 & 90.5 & 75.4 & 90.6 & 85.6 \\
    RoBERTa\textsubscript{base} (\modelshortRidge) & 0.3M & \textbf{87.5} & \underline{94.7} & \underline{91.3} & \textbf{64.2} & 92.6 & \underline{90.6} & 76.5 & \textbf{90.9} & \textbf{86.0} \\
    \midrule

    RoBERTa\textsubscript{large} (FT)* & 355.0M & 90.2 & 96.4 & 90.9 & 68.0 & 94.7 & 92.2 & 86.6 & 92.4 & 88.9 \\
    \midrule
    RoBERTa\textsubscript{large} $(\text{Adapter}^\text{H})$*& 0.8M & \underline{90.3} & \textbf{96.3} & 87.7 & 66.3 & \underline{94.7} & \underline{91.5} & 72.9 & 91.5 & 86.4 \\

    RoBERTa\textsubscript{large} (LoRA) & 0.8M & \textbf{90.7} & \underline{96.2} & \textbf{93.0} & \textbf{68.1} & 94.6 & \textbf{91.6} & \underline{85.2} & \underline{92.0} & \underline{88.9} \\
    RoBERTa\textsubscript{large} $(\text{SVFT}^P)$ & 1.1M & 96.4 & 94.4 & 91.1 & 56.2 & 91.3 & 87.7 & 73.6 & 88.9 & 83.7 \\
    RoBERTa\textsubscript{large} (\modelshortShift) & 0.8M & 89.3 & 95.2 & \underline{92.3} & 66.8 & 93.5 & 90.0 & 84.5 & 91.9 & 88.0 \\
    RoBERTa\textsubscript{large} (\modelshortRidge) & 0.8M & \textbf{90.7} & 95.9 & \textbf{93.0} & \underline{67.8} & \textbf{94.9} & 91.3 & \textbf{86.6} & \textbf{92.2} & \textbf{89.0} \\
    \bottomrule
  \end{tabular}
  \vspace{-2mm}
  \caption{Performance on GLUE tasks. The metrics are Matthews correlation for CoLA, Pearson coefficient for STS-B, F1 for MRPC, and accuracy for other tasks. $\ast$ denotes metrics published in prior works. The values present are averaged over 3 runs on different seeds. Full tuning (FT) statistics are also reported for comparison purposes.}
  \label{tab:NLU_results}
  \vspace{-3mm}
\end{table*}

\begin{table*}[htb]
	\centering
	% \footnotesize
    \small
	\addtolength{\tabcolsep}{-4pt}
	\begin{tabular}{l c ccccc}
		\hline
		\toprule
		\multirow{2}{*}{Method} & \multirow{2}{*}{\# Params} & \multicolumn{5}{c}{E2E} \\
		&  & BLEU & NIST & MET & ROUGE-L & CIDEr \\
		\midrule
		GPT-2\textsubscript{medium} (FT)* & 354.92M & 68.2 & 8.62 & 46.2 & 71.0 & 2.47 \\
            \midrule 
            GPT-2\textsubscript{medium} ($\text{Adapter}^{\text{H}}$)* & 11.09M & 67.3 & 8.50	& 46.0 & 70.7	& 2.44        \\

		% GPT-2\textsubscript{medium} (AutoLoRA)* & 0.3M & 67.9 & 8.68 & 46.0 & 68.9 & 2.37 \\
		GPT-2\textsubscript{medium} (LoRA) & 0.3M & 69.1 & 8.73 & \textbf{46.5} & \textbf{71.4} & \textbf{2.51} \\
		% GPT-2\textsubscript{medium} (LIMTune Shift) & 0.3M & 68.2 & 8.62 & 46.2 & 71.0 & 2.47 \\
            GPT-2\textsubscript{medium} (\modelshortShift) & 0.3M & 69.0 & 8.72 & 46.4 & 70.8 & 2.42 \\
		GPT-2\textsubscript{medium} (\modelshortRidge) & 0.3M & \textbf{69.3} & \textbf{8.75} & 46.3 & 70.8 & \textbf{2.51} \\
		\bottomrule
	\end{tabular}
	\caption{Performance on NLG with beam size as 10. $\ast$ denotes metrics published in prior works. Full tuning (FT) statistics are also reported for comparison purposes. }
    % \AM{\modelshortShift\ values}
	\label{tab:NLG_results}
    \vspace{-5mm}
\end{table*}

\begin{table*}
    \centering
    \small
    \begin{tabular}{lcccccc}
        \toprule
         \multirow{2}{*}{Method} & \multicolumn{3}{c}{Gemma-2B} & \multicolumn{3}{c}{LLaMA-3-8B} \\
         
         \cmidrule(l){2-4} \cmidrule(l){5-7}
         
         & \#Params & GSM-8K & MATH & \#Params & GSM-8K & MATH \\

         \midrule

        FT  & 2.5B & 52.69 & 17.94 & 8.0B & 64.13 & 16.24 \\

        \midrule

        LoRA & 4.9M & 47.23 & \textbf{16.66} & 16.2M & \underline{74.90} & \textbf{25.38} \\

        DoRA & 5.6M & \textbf{51.02} & \underline{16.60} & 17.4M & \textbf{76.70} & \underline{25.10} \\

        ${\text{SVFT}^P}^\ast$ & 0.19M & 40.34 & 14.38 & 0.48M & 69.22 & 20.44 \\

        ${\text{SVFT}_d^R}^\ast$ & 6.35M & \underline{50.03} & 15.56 & 13.1M & 75.90 & 24.22 \\

        \modelshortRidge & 5.67M & 47.38 & 16.38 & 18.8M & 73.99 & 24.04 \\
        
         \bottomrule
    \end{tabular}
    \vspace{-2mm}
    \caption{Accuracy on Mathematical Reasoning (GSM-8K and MATH). $\ast$ denotes metrics published in prior works.}
    \label{tab:math_reasoning}
    \vspace{-5mm}
\end{table*}

We conduct a comprehensive set of experiments across a diverse set of tasks within the domain of Natural Language Understanding and Generation, involving widely used language models with a range of sizes from RoBERTa (base: 125M params, large: 355M params) \cite{liu2019robertarobustlyoptimizedbert} and GPT-2 (medium: 355M params) \cite{Radford2019LanguageMA} to Gemma-2B (2.5B params) \cite{gemmateam2024gemmaopenmodelsbased} and Llama3-8B (8B params) \cite{grattafiori2024llama3herdmodels}. LoRMA has been evaluated against various baselines, including LoRA and its variants like DoRA and SVFT and other PEFT strategies like BitFit and Adapters (\S\ref{sec:related}). We also report the full fine-tuning results for comparison. Overall, LoRMA demonstrates competitive performance to existing approaches.

\subsection{Natural Language Understanding Tasks}
For RoBERTa, we assess the performance of our approach on the GLUE benchmark \cite{wang-etal-2018-glue} on the base and large variants. To maintain consistency with the results and comparison with LoRA, only the query and the value matrices were adapted using our multiplicative techniques. The GLUE benchmark (details in App. \ref{app:dataset-desc}) provides a varied set of tasks ranging from single-sentence tasks (CoLA and SST-2) to similarity and paraphrasing tasks (MRPC, STS-B, QQP) to natural language inference tasks (MNLI, QNLI, RTE). The trends observed in Table \ref{tab:NLU_results} are similar across both variants. On average, LoRMA$_\pi$ performs competitively to LoRA, and LoRMA$_+$ surpasses other PEFT approaches.

\subsection{Natural Language Generation Tasks}
For GPT-2 (medium), we present results on the E2E dataset \cite{novikova-etal-2017-e2e}, commonly used for evaluating NLG capabilities in Table \ref{tab:NLG_results}. Additional GPT-2 experiments, including DART \cite{nan-etal-2021-dart} and WebNLG \cite{gardent-etal-2017-webnlg}, to evaluate our approach have been discussed in App.~\S\ref{app:additional-experiments}. Like RoBERTa, the modified weights included the query and value weights in the spliced \texttt{c\_attn} matrices of GPT-2. As shown in Table \ref{tab:NLG_results}, both \modelshortShift and \modelshortRidge outperform other baselines and are at par with LoRA.

\noindent We further evaluate \modelshortRidge\ on tasks related to mathematical question answering, using larger models like Gemma-2B and LLaMA-3-8B. The pre-trained models were fine-tuned on the MetaMathQA-40K dataset \cite{yu2024metamath}, and then evaluation was done on the GSM-8K \cite{gsm8k} and MATH \cite{hendrycks2021measuringmathematicalproblemsolving} datasets. We adhere to the setup in \citet{svft} for all methods for a fair comparison. The query, key, value, up, down, output, and gate projections (Q, K, V, U, D, O, G) are the weights adapted for Gemma, and the matrices adapted for LLaMA are the up, down, output, and gate projections (U, D, O, G). The results have been presented in Table \ref{tab:math_reasoning}, demonstrating the competitive performance of \modelshortRidge. 

\noindent From the outset, our goal was to introduce an alternative efficient fine-tuning technique, and the overall trends and comparisons across a range of experiments demonstrate that our multiplicative adaptation approach achieves competitive performance relative to several other PEFT methods. However, the main advantage of our approach comes from faster convergence and richer parameter space explored by our approach.

\section{Ablation Studies} \label{sec:results}
\begin{figure*}[htbp]
    \centering
    \includegraphics[scale=0.45]{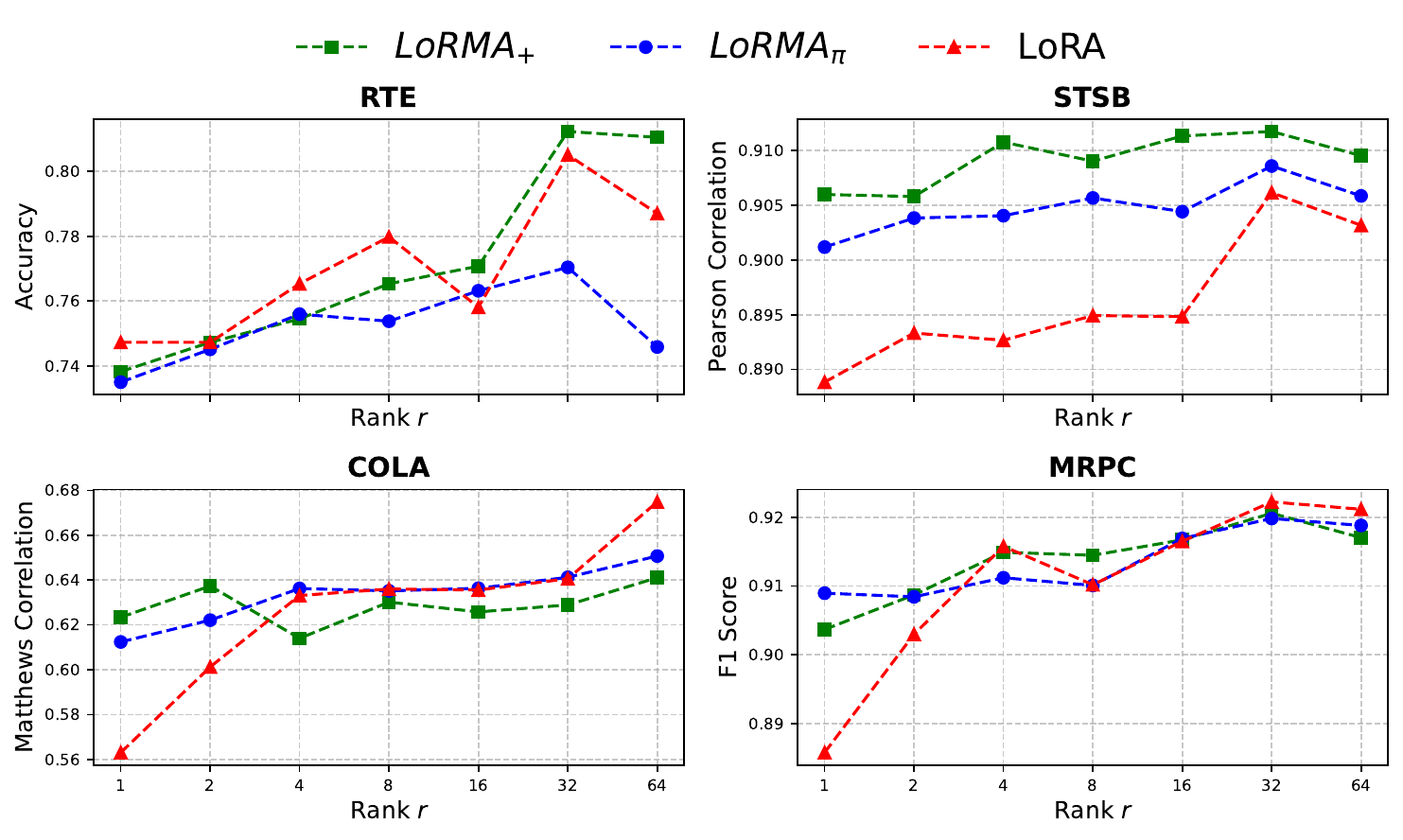}
    \caption{Comparing performance across ranks for the GLUE tasks RTE, STSB, CoLA, MRPC for RoBERTa\textsubscript{base}.}
    \label{fig:rank_comp}
\end{figure*}

%\AM{need to write 7-8 lines describing the results for both, how it compares with previous approaches, etc.}

\label{subsec:faster_conv}
\subsection{Faster convergence of LoRMA}  
\begin{figure}[htbp]
    \centering
    \includegraphics[width=\linewidth]{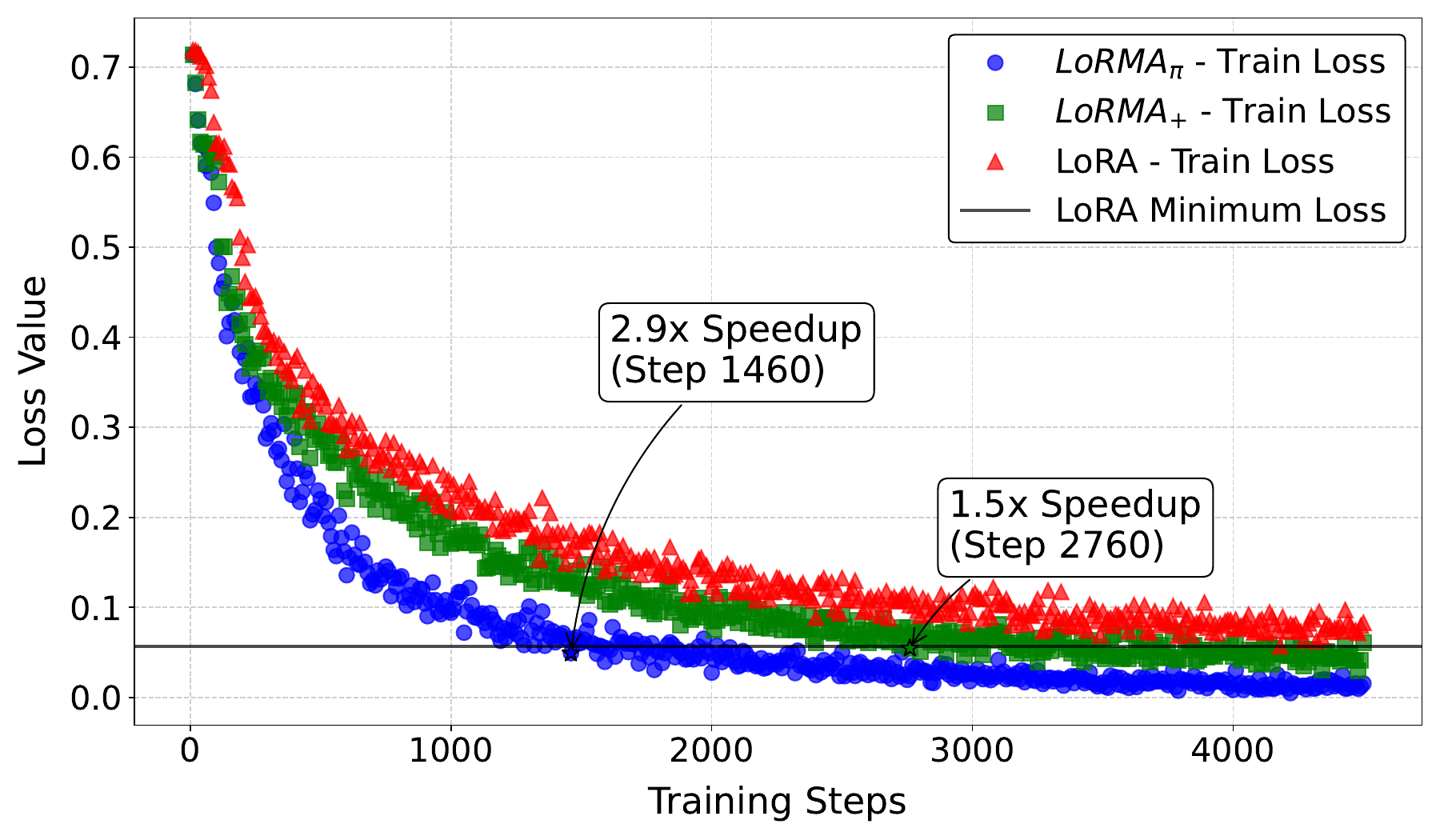}
    \caption{Train loss curves for CoLA: RoBERTa\textsubscript{base} for various techniques.}
    \label{fig:train_loss_cola}
\end{figure}

Convergence time reflects how quickly a model reaches a stable or desirable level of performance during training. To complement the evaluation metrics presented in Table \ref{tab:NLU_results}, we demonstrate in this section that our proposed techniques achieve faster convergence compared to LoRA. We quantify convergence speed using the \textit{Area Under the Curve (AUC)} metric for the training loss curve, where a lower AUC indicates faster convergence. Fig. ~\ref{fig:train_loss_cola} illustrates the training loss curves for LoRMA (both $\mathcal I_+$ and $\mathcal I_\pi$ variants) compared to LoRA on the CoLA task while using RoBERTa\textsubscript{base} model. The results show a steeper decline in training loss. The percentage reduction in AUC for various tasks relative to LoRA is summarized in Table \ref{tab:AUC_conv}. Similar trends were observed for other tasks as well.

\begin{table}[htbp]
  \centering
  \small
  \footnotesize
  \addtolength{\tabcolsep}{-4pt}
  \begin{tabular}{lcc}
  \toprule
  Task & \% AUC $\downarrow \mathcal (I_+)$ & \% AUC $\downarrow \mathcal (I_\pi)$ \\
  \midrule
  SST-2 (RoBERTa\textsubscript{base}) & 10.84 & 30.21 \\ 
  CoLA (RoBERTa\textsubscript{base}) & 23.20 & 51.97 \\ 
  \bottomrule
  \end{tabular}
  \caption{\% AUC decrease in comparison with LoRA}
  \label{tab:AUC_conv}
\end{table}

\subsection{Presence v/s absence of rank-inflation}
% \SP{Can also include rank vs epoch here (may keep/avoid graph)}
% \lipsum[2]
% \noindent\textbf{Presence v/s Absence of Rank-Inflation:} 
\label{subsec:presence-absence-rank} As explained earlier (\S\ref{sec:background}), a naive low-rank multiplicative adaptation of $\mathbf W_0$ has limitations. We present here the empirical verification of the same, and the results are shown in Table~\ref{tab:infl_absence}. The experiments were done on RoBERTa\textsubscript{large} on a subset of GLUE tasks, and all the hyperparameters and training conditions were kept exactly the same, apart from the presence and absence of the rank inflation strategies. The results for the $\mathcal I_+$ have been reproduced for comparison. Further, we evaluate the effectiveness of the proposed rank inflation strategies by monitoring the rank of matrices throughout the training procedure. We observe that these operations successfully help break the rank bottleneck, and the matrices are almost full rank throughout (refer to App.~\S\ref{app:rank-progress}).

\begin{table}[htbp]
  \centering
  \small
  % \footnotesize
  % \addtolength{\tabcolsep}{}
  \begin{tabular}{lcccc}
    \toprule
    Method & MRPC & STS-B & RTE & QQP \\
    \midrule
    LoRMA & 81.2 & 15.6 & 52.7 & 83.0 \\
    \modelshortRidge & \textbf{92.9} & \textbf{92.2} & \textbf{86.6} & \textbf{91.3} \\
    \bottomrule
  \end{tabular}
  \caption{The absence of rank inflation severely limits the model's capabilities. }
  \label{tab:infl_absence}
\end{table}

\begin{table}[htbp]
  \centering
  \small
  % \footnotesize
  % \addtolength{\tabcolsep}{}
  \begin{tabular}{lcccc}
    \toprule
    Method & CoLA & MRPC & STS-B & RTE \\
    \midrule
    \modelshortRidge (\textit{Post}) & \textbf{68.9} & 92.5 & 91.8 & 86.3 \\
    \modelshortRidge (\textit{Pre}) & 67.8 & \textbf{92.9} & \textbf{92.2} & \textbf{86.6} \\
    \bottomrule
  \end{tabular}
  \caption{Pre-multiplication vs Post-multiplication.}
  \label{tab:post_pre}
\end{table}

%\begin{table*}[htbp]

\begin{table*}[t]
\small
  \centering
  \addtolength{\tabcolsep}{4pt}
  \begin{tabular}{l|c|ccccc}
    \toprule
    & Weight Matrix & $r=1$ & $r=2$ & $r=4$ & $r=8$ & $r=64$ \\
    \midrule
    \multirow{3}{*}{MRPC} & $\mathbf W_q$ & 89.6 & 90.5 & 90.2 & 90.2 & 91.2 \\
    & $\mathbf W_q, \mathbf W_k$ & 90.6 & 91.4 & 91.3 & 91.4 & 91.8 \\
    & $\mathbf W_q, \mathbf W_k, \mathbf W_v, \mathbf W_o$ & 90.7 & 91.6 & 91.7 & 91.7 & 93.2 \\
    \midrule
    \multirow{3}{*}{STS-B} & $\mathbf W_q$ & 88.4 & 88.6 & 88.6 & 89.0 & 89.3 \\
    & $\mathbf W_q, \mathbf W_k$ & 89.1 & 89.5 & 89.2 & 89.3 & 89.2 \\
    & $\mathbf W_q, \mathbf W_k, \mathbf W_v, \mathbf W_o$ & 91.0 & 91.2 & 90.9 & 90.9 & 91.1  \\
    \bottomrule
  \end{tabular}
  \caption{RoBERTa\textsubscript{base} with \modelshortRidge. Validation accuracy across different weights being adapted with varying ranks $r$ for the GLUE tasks MRPC and STS-B.}
  \label{tab:rank_v_perf}
\end{table*}

\begin{table*}[h!]
\footnotesize
  \centering
  \addtolength{\tabcolsep}{0pt}
\begin{tabular}{l|lll|lll}
\toprule
 %& \multicolumn{6}{c}{RoBERTa-Large} \\
 %\midrule
 & \multicolumn{3}{c}{Layer 3} & \multicolumn{3}{c}{Layer 23} \\
 \midrule
 \textbf{Metric} & $\Delta \mathbf W_\text{\modelshortRidge}$ & $\Delta \mathbf W_\text{\modelshortShift}$ & Random  & $\Delta \mathbf W_\text{\modelshortRidge}$ & $\Delta \mathbf W_\text{\modelshortShift}$ & Random \\
 \midrule
 $(\downarrow)\ \left\Vert  W - \Delta\mathbf W_{\text{LoRA}} \right\Vert_F$ & 3.54 & 31.93 & 1024.27 & 10.02  & 38.31 & 1022.40  \\
 $(\uparrow)\ \texttt{cos} ( W, \Delta\mathbf W_{\text{LoRA}})$ & $74.2 \times 10^{-3}$ & $4.1\times 10^{-3}$ & $0.1 \times 10{-3}$ & $68.1\times10^{-3}$ & $0.3\times10^{-3}$  & $-1.2\times10^{-3}$   \\
$(\downarrow)\ \left(  W, \Delta\mathbf W_{\text{LoRA}} \right)_{\mathcal S}^r$ & 0.99 & 8.93 & 176.18 & 3.75 & 13.40  & 175.67\\
$(\downarrow)\ \left( W, \Delta\mathbf W_{\text{LoRA}} \right)_{\mathcal E}^r$ & 0.06 & 2.67 & 89.07 & 0.49 & 3.31  & 89.70\\
$(\downarrow)\ \Theta_1(W,\Delta\mathbf W_{\text{LoRA}})$ & $2.28\times 10^{-6}$ & $2.27\times 10^{-6}$ & 1.56 & $2.34\times 10^{-6}$ & $2.32\times 10^{-6}$ & 1.57 \\
\bottomrule
\end{tabular}
\caption{Correlation between $\Delta\mathbf W_\text{LoRA}$ and $ \Delta \mathbf W_\text{LoRMA}$ for RoBERTa\textsubscript{large}. {\small $\uparrow/\downarrow$ indicates higher / lower is more similar.}}
\label{tab:weight_analysis}
\end{table*}

\begin{table*}[h!]
  \small
  \centering
  \addtolength{\tabcolsep}{4pt}
  \begin{tabular}{l|cccccc}
    \toprule
    & \multicolumn{6}{c}{\# Trainable Parameters $\approx$ 150K} \\ 
    \midrule
    Weight Matrix & $\mathbf W_q$ & $\mathbf W_k$ & $\mathbf W_v$ & $\mathbf W_q,\mathbf W_v$ & $\mathbf W_q, \mathbf W_k$ & $\mathbf W_q, \mathbf W_k, \mathbf W_v, \mathbf W_o$ \\
    $r$ & 8 & 8 & 8 & 4 & 4 & 2 \\
    \midrule
    MRPC & 90.2 & 91.0 & 91.4 & 90.2 & 91.3 & \textbf{91.6} \\
    STS-B & 89.0 & 89.3 & 90.9 & 90.5 & 89.2 & \textbf{91.2} \\
    \bottomrule
  \end{tabular}
  \caption{RoBERTa\textsubscript{base} with \modelshortRidge on a fixed budget for the GLUE tasks MRPC and STS-B, with scaling factor $\alpha = r$ for respective $r$'s depending upon the application.}
  \label{tab:fix_budget}
  \vspace{-5mm}
\end{table*}
\subsection{Pre-multiplication v/s Post multiplication}
% \noindent\textbf{Pre-multiplication v/s Post multiplication:}
The Corollary~\ref{cor:square_mult} allows for an equivalent representation of the multiplicative transformation for square matrices, i.e., post and pre-multiplication. We test \textit{post-multiplicative} \modelshortRidge\ (Table~\ref{tab:post_pre}) and observe almost comparable performance with the strategy mentioned above. %This verifies the discussion in \S\ref{sec:background}.

\subsection{Choice of Weight Matrix}
% \noindent\textbf{Choice of Weight Matrix:}
With a fixed parameter budget, it becomes crucial to strategically allocate adaptive weights to achieve optimal performance. To investigate this, we set a parameter budget of approximately 150K parameters, corresponding to $r=8$ for single-weight-type adaptation across the GLUE tasks MRPC and STS-B. The adapted model is RoBERTa\textsubscript{base}, with a scaling factor $\alpha = r$ (for varying ranks $r$) used in the additive variant of our method (\modelshortRidge). The trends observed in Table~\ref{tab:fix_budget} suggest that, given a fixed budget, diversifying the adaptive tuning---i.e., distributing the adaptation across multiple weight matrices---leads to better performance.

\subsection{Rank v/s Performance}
To see the effect of rank $r$ over performance, we adapt $\mathbf W_q$, $\{ \mathbf W_q, \mathbf W_k \}$ and $\{ \mathbf W_q, \mathbf W_k, \mathbf W_v, \mathbf W_o \}$ weight matrices with \modelshortRidge and the results are compiled in Table \ref{tab:rank_v_perf}. This observation aligns with similar experiments conducted on \modelshortRidge, \modelshortShift, and LoRA (Fig.~\ref{fig:rank_comp}). The overarching trend shows that performance improves with higher ranks across all techniques. However, this trend is neither strict nor monotonic, as performance dips at higher ranks are also observed. This could possibly be due to a low intrinsic rank of $\Delta \mathbf W$ being sufficient to capture the transformation and higher ranks leading to over-parametrization rather than learning additional information. Notably, LoRMA scales effectively across different ranks and demonstrates comparable or even superior performance to LoRA, particularly in highly parameter-constrained scenarios. This underscores the scalability and effectiveness of LoRMA, along with its rank-inflation variants, in resource-constrained settings.

\subsection{Comparison with $\Delta\mathbf W_\text{LoRA}$}
% \noindent\textbf{Comparision with $\Delta\mathbf W_\text{LoRA}$:} 
\label{subsec:weight-update-comparision} 
For any technique, denote $\Delta \mathbf W$ to be the difference between the final adapted weight matrix and the initial weight matrix (the frozen weights). We investigate the relationship of $\Delta\mathbf{W}_{\text{LoRA}}$ with $\Delta\mathbf{W}_{\text{\modelshortRidge}}$ and $\Delta\mathbf{W}_{\text{\modelshortShift}}$ as compared to a random matrix. To assess the correlation, we employ a variety of metrics, the results of which are summarized in Table~\ref{tab:weight_analysis}. We utilize the Frobenius norm ($\left\Vert\cdot\right\Vert_F$) to measure the deviation between the matrices. The cosine similarity of the flattened matrices ($\texttt{cos}(\cdot, \cdot)$) and principal subspace angle $\Theta_1(\cdot,\cdot)$ between their column spaces has been used to measure their alignment. We compute the sum of squared differences between the top-$r$ singular values $(\cdot, \cdot)_\mathcal S^r$ and eigenvalues $(\cdot, \cdot)_\mathcal E^r$ of the two matrices to assess their similarity. As can be seen in Table \ref{tab:weight_analysis}, the main trend points towards a high correlation between $\Delta\mathbf W_\text{LoRA}$ and $\Delta \mathbf W_\text{\modelshortRidge}$ and $\Delta \mathbf W_\text{\modelshortShift}$, which shows that our multiplicative techniques can capture updates learned by additive LoRA. 
To assess the expressibility of the transformations, we compare the rank of $\Delta \mathbf W$. For LoRA, $\Delta \mathbf W = \mathbf B \mathbf A$; hence, it is restricted to be a low-rank update (property \ref{eqn:product}). While for \modelshortShift, there are no such limitations. We empirically observe them to be almost full-rank matrices (refer to App. \S\ref{app:rank-of-weight-update}).

\subsection{LoRMA$_+$ vs LoRMA$_\pi$}
The usage of different rank inflation strategies makes the parameter exploration space of \modelshortShift\ different compared to \modelshortRidge, which leads to differences in impact. While the convergence rate of both multiplicative methods is higher than that of additive LoRA, as shown in Fig. \ref{fig:train_loss_cola}, convergence is faster for \modelshortShift\ than \modelshortRidge. Overall, the performance of \modelshortShift, though comparable, is seen to be slightly lower than \modelshortRidge. We analyze this in App. \S\ref{app:pi-vs-plus}. We find LoRMA$_\pi$ to demonstrate better, similar, or slightly lower capability to learn compared to \modelshortRidge, based on the train-eval loss curves. Choosing the better of the two approaches is task-specific.
\section{Conclusion} \label{sec:conclusion}

In this work, we proposed LoRMA, a novel approach for updating the weights of a language model via multiplicative updates. We mathematically proved the existence of multiplicative updates. Further, to overcome the limitations of the naive approach of multiplicative updates, we propose two methods to inflate the rank of the update matrix via permutation and additive identity. Extensive experiments demonstrate the competitive performance and training efficacy of the proposed approach. In the future, we plan to experiment with combining LoRMA with existing LoRA-based enhancements like AutoLoRA, DyLoRA, etc. 

\section*{Limitations}
\label{sec:limitations}

%\textbf{Guarantee of full rank via inflation.} In the paper, two rank inflation strategies have been proposed, which are the additive $\mathcal{I}_+$ and permutation-based rank inflation strategies $\mathcal{I}_{\pi}$. For $\mathcal{I}_+$ with the help of property \ref{eqn:sum}, the existence of a guaranteed lower bound of $d -r$ on the rank was shown. While these strategies significantly assist in inflating the rank, it might be possible while they are almost full rank but not full rank. 

\noindent\textbf{The ability to plug out the parameters.} In a production setting, LoRA converts a base model to the model tuned to a task by adding $BA$ to the weight matrix of the model, and one can recover the original model by subtracting out the original weight matrix. In the case of LoRMA, by updating the original weight matrix by multiplication with $\mathcal{I}(BA)$, the tuned model can be deployed. The recovery of original model weights from the updated form would require $\mathcal{I}(BA)$ to be invertible, which might not be the case, as discussed above. To mitigate this, a copy of the original parameters would have to be maintained.

\noindent\textbf{Time complexity of $\mathcal{I}_{\pi}$ during training.} As discussed in Section \ref{sec:methods}, while other variants of LoRMA have a similar order of time complexity as LoRA during the training process, \modelshortShift\ has a slightly higher time complexity at training time. However, by merging weights during inference time, all of them would have no inference latency, which makes the method still a viable option.

% \SP{Is below applicable, if yes then keep else we may remove.}

% \noindent\textbf{Experiments with Smaller Models.} In this paper, in order to be comparable with previous works, we experimented mainly with RoBERTa and GPT-2 models. We performed experiments on a variety of tasks, and the results are indicative of the efficacy of the proposed method. We expect the results to be generalized to larger LLMs as well. 

\section*{Ethical Considerations}
\label{sec:ethical-considerations}

We abide by the ACL Code of Ethics code during our research. This work introduces a new variant of parameter-efficient fine-tuning approaches for LLMs that do not directly have possible harms associated with them. The use of LLMs has ethical considerations that should be kept in mind. We have used public models (RoBERTa, GPT2, Gemma, and LLaMA) and public datasets (GLUE, E2E, WebNLG, DART, MetaMath40K, GSM-8K, MATH) to evaluate the effectiveness of our proposed approach. 

% \AM{update list of models and tasks}
%We provide the details of the computing resources and hyper-parameters for reproducibility. %We plan to release our code and models after the acceptance of the paper. 

% % Bibliography entries for the entire Anthology, followed by custom entries
% %\bibliography{anthology,custom}
% % Custom bibliography entries only
% \bibliography{custom}
\bibliography{references}

\clearpage
\newpage

\appendix

\section*{Appendix}
\appendix

%%%%%%%%%%%%%%%%%%%%%%%%%%%

\titlecontents{section}[18pt]{\vspace{0.05em}}{\contentslabel{1.5em}}{}
{\titlerule*[0.5pc]{.}\contentspage} % Set the formatting for appendix sections in the table of contents

% % for list of tables
% \titlecontents{table}[0pt]{\vspace{0.05em}}{\contentslabel{1em}}{}
% {\titlerule*[0.5pc]{.}\contentspage} % Set the formatting for appendix tables in the list of tables

% for list of figures
\titlecontents{table}[0pt]{\vspace{0.05em}}{\contentslabel{1em}}{}
{\titlerule*[0.5pc]{.}\contentspage} % Set the formatting for appendix tables in the list of tables

\startcontents[appendix] % Start the table of contents for the appendix
\section*{Table of Contents} % Title for the appendix table of contents
%\addcontentsline{toc}{section}{Table of Contents} % Add the appendix table of contents to the main table of contents
\printcontents[appendix]{section}{0}{\setcounter{tocdepth}{4}} % Print the table of contents for the appendix

\startlist[appendix]{lot} % Start the list of tables for the appendix
\section*{List of Tables} % Title for the appendix list of tables
%\addcontentsline{lot}{section}{List of Tables} % Add the appendix list of tables to the main list of tables
\printlist[appendix]{lot}{}{\setcounter{tocdepth}{1}} % Print the list of tables for the appendix

\startlist[appendix]{lof} % Start the list of tables for the appendix
\section*{List of Figures} % Title for the appendix list of tables
%\addcontentsline{lot}{section}{List of Tables} % Add the appendix list of tables to the main list of tables
\printlist[appendix]{lof}{}{\setcounter{tocdepth}{1}} % Print the list of tables for the appendix

\clearpage
\newpage

%%%%%%%%%%%%%%%%%%%%%%%%%%%

%\section{iSign Benchmark Repository}

%%%%%%%%%%%%%%%%%%%%%%%%%%%
\section{Time complexity calculations} \label{app:time-complexity}
Here, we describe the strategic re-ordering of operations to mitigate the large time complexity incurred due to matrix multiplications. These have been summarized in Table \ref{tab:time-complexity-calculation}.
% LoRA
\subsection*{LoRA}
Multiply $\mathbf W_0$ with $\mathbf x$ ($\mathcal{O}(dkb)$). Multiply $\mathbf A$ with $\mathbf x$ ($\mathcal{O}(krb)$). Multiply $\mathbf B$ with $\mathbf A \mathbf x$ ($\mathcal{O}(drb)$). Add $\mathbf W_0 \mathbf x$ with $\mathbf B \mathbf A \mathbf x$ ($\mathcal{O}(db)$).
% LoRMA
\subsection*{LoRMA}
Multiply $\mathbf W_0$ with $\mathbf x$ ($\mathcal{O}(dkb)$). Multiply $\mathbf A$ with $\mathbf W_0 \mathbf x$ ($\mathcal{O}(drb)$). Multiply $\mathbf B$ with $\mathbf A \mathbf W_0 \mathbf x$ ($\mathcal{O}(drb)$).
% LoRMA Pi
\subsection*{\modelshortShift}
Multiply $\mathbf W_0$ with $\mathbf x$ ($\mathcal{O}(dkb)$). Multiply $\mathbf B$ with $\mathbf A$  ($\mathcal{O}(dkr)$). Inflation $\mathcal{I}_{\pi}$ of $BA$ ($\mathcal{O}(d^2r)$). Multiply $\mathcal{I}_{\pi}(BA)$) with $\mathbf W_0 \mathbf x$ ($\mathcal{O}(d^2b)$).
% LoRMA +
\subsection*{\modelshortRidge}
Multiply $\mathbf W_0$ with $\mathbf x$ ($\mathcal{O}(dkb)$) for first term. Multiply $\mathbf W_0$ with $\mathbf x$ ($\mathcal{O}(dkb)$) for second term. Multiply $\mathbf A$ with $\mathbf W_0 \mathbf x$ ($\mathcal{O}(drb)$). Multiply $\mathbf B$ with $\mathbf A \mathbf W_0 \mathbf x$ ($\mathcal{O}(drb)$). Add $\mathbf W_0 \mathbf x$ with $\mathbf B \mathbf A \mathbf W_0 \mathbf x$ ($\mathcal{O}(db)$).
\begin{table}[htb]
\addtolength{\tabcolsep}{-2pt}
\small
\begin{tabular*}{\linewidth}{ccc}
\toprule
Method & Computation & Complexity Calculation \\
\midrule
\multirow{2}{*}{LoRA} & \multirow{2}{*}{$(\mathbf W_0 + \mathbf{BA})\mathbf{x}$} & $dkb + krb + drb + db$ \\
                  &                   & $\mathcal{O}(dkb)$ \\
\midrule
\multirow{2}{*}{LoRMA} & \multirow{2}{*}{$\mathbf{BAW}_0\mathbf{x}$} & $dkb + 2drb$ \\
                  &                   & $\mathcal{O}(dkb)$ \\
\midrule
\multirow{2}{*}{\modelshortShift} & \multirow{2}{*}{$\mathcal{I}_{\pi}(\mathbf{BA}) \mathbf W_0\mathbf x$} & $dkb + dkr + d^2r + d^2b$ \\
                  &                   & $\mathcal{O}(d^2(r + b))$ \\
\midrule
\multirow{2}{*}{\modelshortRidge} & \multirow{2}{*}{$\mathbf W_0\mathbf{x} + \mathbf{BA} \mathbf W_0\mathbf x$} & $2dkb + 2drb + db$ \\
                  &                   & $\mathcal{O}(dkb)$ \\
\bottomrule
\end{tabular*}
\caption{Time Complexity for computations incurred by different methods during training time.}
\label{tab:time-complexity-calculation}
\end{table}

\section{Dataset description} \label{app:dataset-desc}

\begin{itemize}
    \item \textbf{GLUE Benchmark}: The benchmark comprises wide-ranging natural language understanding tasks mostly restricted to English. It consists of tasks like CoLA (\cite{colawarstadt2018neural}, grammatical acceptability), SST-2 (\cite{sst2-socher-etal-2013-recursive}, sentiment analysis), MRPC (\cite{mrpcdolan-brockett-2005-automatically}, semantic textual similarity), STS-B (\cite{stsbCer_2017}, semantic textual similarity), QQP (\cite{qqpsharma2019naturallanguageunderstandingquora}, question answers), and inference tasks like MNLI \cite{mnliwilliams2018broadcoveragechallengecorpussentence}, QNLI \cite{qnlirajpurkar2018knowdontknowunanswerable} as well as RTE \cite{rtepoliak-2020-survey}.

    The datasets are available under public license and were used using the \texttt{datasets} API provided by HuggingFace. The dataset statistics are presented in Table~\ref{tab:dataset_stats}.

    \item \textbf{E2E NLG Challenge}: The E2E dataset was released in \citet{novikova-etal-2017-e2e} and is a popular dataset for testing efficacy in natural language generation tasks. The dataset consists of 42061 training samples, 4672 dev, and 4693 test samples. Success on this task is typically measured by achieving high BLEU, NIST, METEOR, Rouge-L, and CIDEr scores, as presented in the paper.

    \item \textbf{DART}: This is a large dataset for open-domain record-to-text generation published in \cite{nan-etal-2021-dart}. It has a total of close to 82K samples. The underlying task is \textit{rdf-to-text} (mapping entity relations to text). 

    \item \textbf{WebNLG Challenge}: WebNLG challenge \cite{gardent-etal-2017-webnlg} is yet another dataset that consists of mapping data to text. Data is a set of triples, and text is the verbalization of this data. It has close to 22K total samples. It involves examples from 9 distinct \href{https://www.dbpedia.org/}{DBPedia} categories during training, with the complete dataset having 15 categories.

    \item \textbf{GSM-8K}: GSM-8K \cite{gsm8k} is a dataset of 8.5K  grade school math word problems created by human problem writers. The dataset has 7.5K training and 1K test problems, and solutions primarily involve basic arithmetic \href{https://paperswithcode.com/dataset/gsm8k}{[reference]}.

    \item \textbf{MATH}: MATH \cite{hendrycks2021measuringmathematicalproblemsolving} is a dataset of 12,500 challenging competition mathematics problems. Each problem in MATH has a full step-by-step solution that can be used to teach models to generate answer derivations and explanations \href{https://paperswithcode.com/dataset/math}{[reference]}.

\end{itemize}

\begin{table*}[htbp]
  \centering
  \small
  \addtolength{\tabcolsep}{4pt}
  \begin{tabular}{l|cccccccc}
    \toprule
    & CoLA & SST-2 & MRPC & STS-B & QQP & MNLI & QNLI & RTE \\
    \midrule
    Train & 8551 & 67349 & 3668 & 5749 & 363871 & 392702 & 104743 & 2490 \\
    Validation & 1043 & 872 & 408 & 1500 & 40432 & 9815 & 5463 & 277 \\
    \bottomrule
    % Train & 8551 & 
  \end{tabular}
  \caption{GLUE Benchmark statistics}
  \label{tab:dataset_stats}
\end{table*}

\section{Additional Experiments} \label{app:additional-experiments}

We conduct more experiments to test and benchmark the capabilities of our multiplicative adapters. In this series, we repeat our NLG experiments for DART \cite{nan-etal-2021-dart} and WebNLG challenge \cite{gardent-etal-2017-webnlg}. The trends are similar to that observed in Table~\ref{tab:NLG_results}. Our adapters perform comparably in evaluation with LoRA. The results are presented in Table~\ref{tab:extra-NLG_results}.

\begin{table*}[htb]
	\centering
	% \footnotesize
    \small
	\addtolength{\tabcolsep}{-2pt}
	\begin{tabular}{l c ccccc|ccc|ccc}
		\hline
		\toprule
		Method & \# Params & \multicolumn{5}{c}{E2E} & \multicolumn{3}{c}{DART} & \multicolumn{3}{c}{WebNLG} \\
		& (M) & BLEU & NIST & MET & ROUGE-L & CIDEr & BLEU & MET & TER & BLEU & MET & TER \\
		\midrule
		& & \multicolumn{5}{c}{Beam size 15} & \multicolumn{3}{c}{Beam size 10} & \multicolumn{3}{c}{Beam size 10} \\
		\midrule
		 LoRA & 0.3M & 67.5 & 8.53 & \textbf{46.2} & \textbf{70.8} & 2.49 & \textbf{45.35} & \textbf{0.38} & \textbf{0.53} & \textbf{52.27} & \textbf{0.40} & \textbf{0.45} \\
		\modelshortRidge & 0.3M & \textbf{68.4} & \textbf{8.63} & 46.1 & 70.6 & \textbf{2.50} & 43.64 & \textbf{0.38} & \textbf{0.53} & 49.98 & 0.38 & 0.47 \\
		\bottomrule
	\end{tabular}
	\caption{Extra results for GPT-2\textsubscript{medium} for NLG. For all the metrics, higher is better except TER.}
	\label{tab:extra-NLG_results}
\end{table*}

\section{Hyperparameters and Training Setup} \label{app:hyperparam}
\noindent  We adhere to the standard experimental setup used in LoRA to ensure consistency with prior work. Specifically, our multiplicative transformation technique is applied to the query ($W_{q}$) and value ($W_{v}$) matrices within the attention mechanism of the models. This means that for a 12-layer RoBERTa\textsubscript{base} or GPT-2 M model, our multiplicative adapters are applied a total of 24 times—once for each query and value matrix across all layers. Similarly, for the 24-layer RoBERTa\textsubscript{large} model, the multiplicative adapter is applied 48 times. We use the pre-trained versions of RoBERTa\textsubscript{base} (125M parameters) and RoBERTa\textsubscript{large} (355M parameters) available in the \href{https://huggingface.co/docs/transformers/index}{HuggingFace Transformers library}  \cite{wolf2020huggingface}. We employed the \href{https://huggingface.co/docs/peft/index}{PEFT} \cite{pefT} support on the HuggingFace where available for running experiments. For NLG experiments based on GPT-2, we draw inspiration from \href{https://github.com/microsoft/LoRA}{LoRA's published code}. The pre-trained GPT-2 models have been made available by HuggingFace.
% \AM{also describe the GPT parameters and library}

% \AM{need to write the values of alpha and r in \modelshortRidge} 
\subsection{RoBERTa}
\label{app:hps_roberta}
We utilize AdamW optimizer \cite{loshchilov2018decoupled} along with a linear learning rate decay schedule. The results reported are the mean of runs for three random seeds, with the result for a single run taken to be from the best epoch. The pre-trained RoBERTa model is taken and fine-tuned for each task separately. The hyperparameters have been presented in Table \ref{tab:hyper_roberta}. For the results of previous works, refer to \citet{zaken2021bitfit, houlsby2019parameterefficienttransferlearningnlp, zhang-etal-2024-autolora}. To maintain consistency RoBERTa (both base and large variants) were adapted on only the query and value matrices within the model.

\begin{table*}[h]
    \small
    \addtolength{\tabcolsep}{-1pt}
    \centering
    \begin{tabular}{ll|cccccccc}
        \hline
        \toprule
        Model and method  & Task     & MNLI & SST-2 & MRPC & CoLA & QNLI & QQP & RTE & STS-B \\
        \midrule
                              & Optimizer   & \multicolumn{8}{c}{AdamW} \\
                              & Warmup Ratio & \multicolumn{8}{c}{0.06} \\
                              & LR Schedule & \multicolumn{8}{c}{Linear} \\
        \midrule
        \multirow{5}{*}{\makecell{RoBERTa\textsubscript{base} \\ \modelshortRidge}} \ 
                              & Batch Size & 64 & 64 & 32 & 64 & 64 & 16 & 32 & 16 \\
                              & Epochs & 30 & 60 & 30 & 100 & 25 & 25 & 80 & 40 \\
                              & Learning Rate & 4E-4 & 5E-4 & 4E-4 & 4E-4 & 4E-4 & 5E-4 & 5E-4 & 4E-4 \\
                              & Matrices and $r$ & \multicolumn{8}{c}{$r_q=r_v=8$} \\
                              & Scaling $\alpha$ & 4 & 8 & 8 & 4 & 4 & 8 & 8 & 8 \\
                              & Max Seq. Len. & \multicolumn{8}{c}{512} \\
                              & Weight decay & 0.1 & 0.1 & 0.1 & 0.2 & 0.1 & 0.1 & 0.1 & 0.1 \\
        \midrule
        \multirow{5}{*}{\makecell{RoBERTa\textsubscript{base} \\ \modelshortShift}} \ 
                              & Batch Size & 64 & 64 & 32 & 64 & 64 & 16 & 32 & 16 \\
                              & Epochs & 30 & 60 & 30 & 100 & 25 & 25 & 80 & 40 \\
                              & Learning Rate & 4E-4 & 5E-4 & 4E-4 & 4E-4 & 4E-4 & 5E-4 & 5E-4 & 4E-4 \\
                              & Matrices and $r$ & \multicolumn{8}{c}{$r_q=r_v=8$} \\
                              & Scaling $\alpha$ & \multicolumn{8}{c}{8} \\
                              & Max Seq. Len. & \multicolumn{8}{c}{512} \\
                              & Weight decay & \multicolumn{8}{c}{0.1} \\
        \midrule
        \multirow{5}{*}{\makecell{RoBERTa\textsubscript{base} \\ LoRA}} \ 
                              & Batch Size & 64 & 64 & 32 & 64 & 64 & 16 & 32 & 16 \\
                              & Epochs & 30 & 60 & 30 & 100 & 25 & 25 & 80 & 40 \\
                              & Learning Rate & 4E-4 & 5E-4 & 4E-4 & 4E-4 & 4E-4 & 5E-4 & 5E-4 & 4E-4 \\
                              & Matrices and $r$ & \multicolumn{8}{c}{$r_q=r_v=8$} \\
                              & Scaling $\alpha$ & \multicolumn{8}{c}{8} \\
                              & Max Seq. Len. & \multicolumn{8}{c}{512} \\
                              & Weight decay & \multicolumn{8}{c}{0.1}  \\
        \midrule
        \multirow{5}{*}{\makecell{RoBERTa\textsubscript{large} \\ \modelshortRidge}} \ 
                              & Batch Size & 8 & 8 & 4 & 8 & 4 & 4 & 8 & 4 \\
                              & Epochs & 10 & 10 & 20 & 20 & 20 & 20 & 20 & 10 \\
                              & Learning Rate & 3E-4 & 4E-4 & 3E-4 & 3E-4 & 2E-4 & 3E-4 & 4E-4 & 3E-4 \\
                              & Matrices and $r$ & \multicolumn{8}{c}{$r_q=r_v=8$} \\
                              & Scaling $\alpha$ & 8 & 8 & 4 & 4 & 4 & 8 & 4 & 4 \\
                              & Max Seq. Len. & 128 & 512 & 512 & 128 & 512 & 512 & 512 & 128 \\
                              & Weight decay & 0.1 & 0.1 & 0.1 & 0.2 & 0.1 & 0.1 & 0.1 & 0.1 \\
        \midrule
        \multirow{5}{*}{\makecell{RoBERTa\textsubscript{large} \\ \modelshortShift}} \ 
                              & Batch Size & 8 & 8 & 4 & 8 & 4 & 4 & 8 & 4 \\
                              & Epochs & 10 & 10 & 20 & 20 & 20 & 20 & 20 & 10 \\
                              & Learning Rate & 3E-4 & 4E-4 & 3E-4 & 3E-4 & 2E-4 & 3E-4 & 4E-4 & 3E-4 \\
                              & Matrices and $r$ & \multicolumn{8}{c}{$r_q=r_v=8$} \\
                              & Scaling $\alpha$ & \multicolumn{8}{c}{8} \\
                              & Max Seq. Len. & 128 & 512 & 512 & 128 & 512 & 512 & 512 & 128 \\
                              & Weight decay & \multicolumn{8}{c}{0.1} \\
        \midrule
        \multirow{5}{*}{\makecell{RoBERTa\textsubscript{large} \\ LoRA}} \ 
                              & Batch Size & 8 & 8 & 4 & 8 & 4 & 4 & 8 & 4 \\
                              & Epochs & 10 & 10 & 20 & 20 & 20 & 20 & 20 & 10 \\
                              & Learning Rate & 3E-4 & 4E-4 & 3E-4 & 3E-4 & 2E-4 & 3E-4 & 4E-4 & 3E-4 \\
                              & Matrices and $r$ & \multicolumn{8}{c}{$r_q=r_v=8$} \\
                              & Scaling $\alpha$ & \multicolumn{8}{c}{8} \\
                              & Max Seq. Len. & 128 & 512 & 512 & 128 & 512 & 512 & 512 & 128 \\
                              & Weight decay & \multicolumn{8}{c}{0.1} \\

        \bottomrule
    \end{tabular}
    \caption{The hyperparameters used for RoBERTa on the GLUE benchmark.}
    \label{tab:hyper_roberta}
\end{table*}

\subsection{GPT-2}
\label{app:hps_gpt2}
The GPT-2 models have been trained via the AdamW optimizer using a linear learning rate schedule for 5 epochs. Table \ref{tab:hyper_gpt2} presents the hyper-parameters used for the experiments. For the results of previous works, refer to \citet{zhang-etal-2024-autolora, hu2022lora}. For all GPT-2 experiments, much like RoBERTa, the model was adapted only for the query and value matrices (the spliced \texttt{c\_attn}).

\subsection{Gemma}

Gemma-2B was trained using the AdamW optimizer and a cosine scheduler. The matrices adapted for Gemma are the query, key, value, up, down, output, and gate projections (Q, K, V, U, D, O, G). The remaining hyperparameters are presented in Table \ref{tab:math_hyperparam}. For $\text{SVFT}^R_d$, $d=16$ for Gemma family. For LoRMA, $r$ is set to 4 to maintain a comparable number of training parameters.

\subsection{Llama}

LLaMA-3-8B was trained using the AdamW optimizer and a cosine scheduler. The matrices adapted for LLaMA are the up, down, output, and gate projections (U, D, O, G). The remaining hyperparameters are presented in Table \ref{tab:math_hyperparam}. For $\text{SVFT}^R_d$, $d=12$ for LLaMA family. or LoRMA, $r$ is set to 8 to maintain a comparable number of training parameters.

\subsection{Parameter count in \modelshortRidge\ in LLaMA and Gemma Models}
In LoRMA where $\mathbf{W} = \mathbf{BAW_0}$, to maintain consistency of matrix dimensions, $\mathbf{B} \in \mathbb{R}^{d \times r}, \mathbf{A} \in \mathbb{R} ^ {r \times d}$. While in LoRA $\mathbf{B} \in \mathbb{R}^{d \times r}, \mathbf{A} \in \mathbb{R} ^ {r \times k}$, since $\mathbf{W} = \mathbf{W_0} + \mathbf{BA}$. For models in which the adapted matrices are square, like RoBERTa, GPT-2, and their larger variants, with $d=k$, the number of parameters in LoRMA is the same as those in LoRA. If it is the case that $d > k$, then the number of parameters in LoRMA is slightly higher than LoRA, as can be seen for Gemma and Llama.

\section{Ablation}
\subsection{Performance of LoRMA$_\pi$ vs LoRMA$_+$} \label{app:pi-vs-plus}
\begin{figure*}[h]
    \centering
    \begin{subfigure}[c]{\textwidth}
        \centering
        \includegraphics[width=0.6\textwidth]{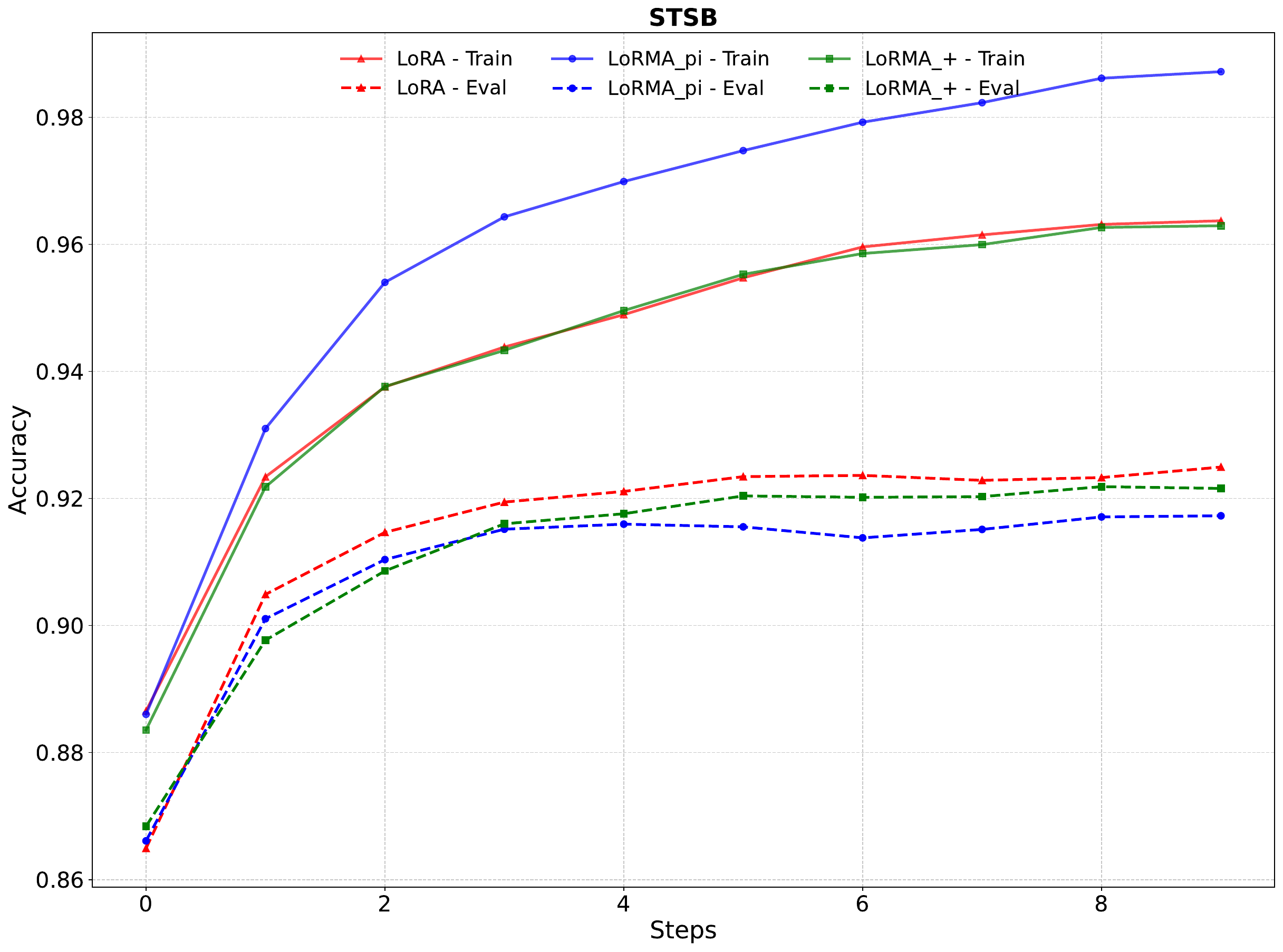}
        \caption{LoRMA$_\pi$ overfitting.}
    \end{subfigure}
    
    \begin{subfigure}[c]{\textwidth}
        \centering
        \includegraphics[width=0.6\textwidth]{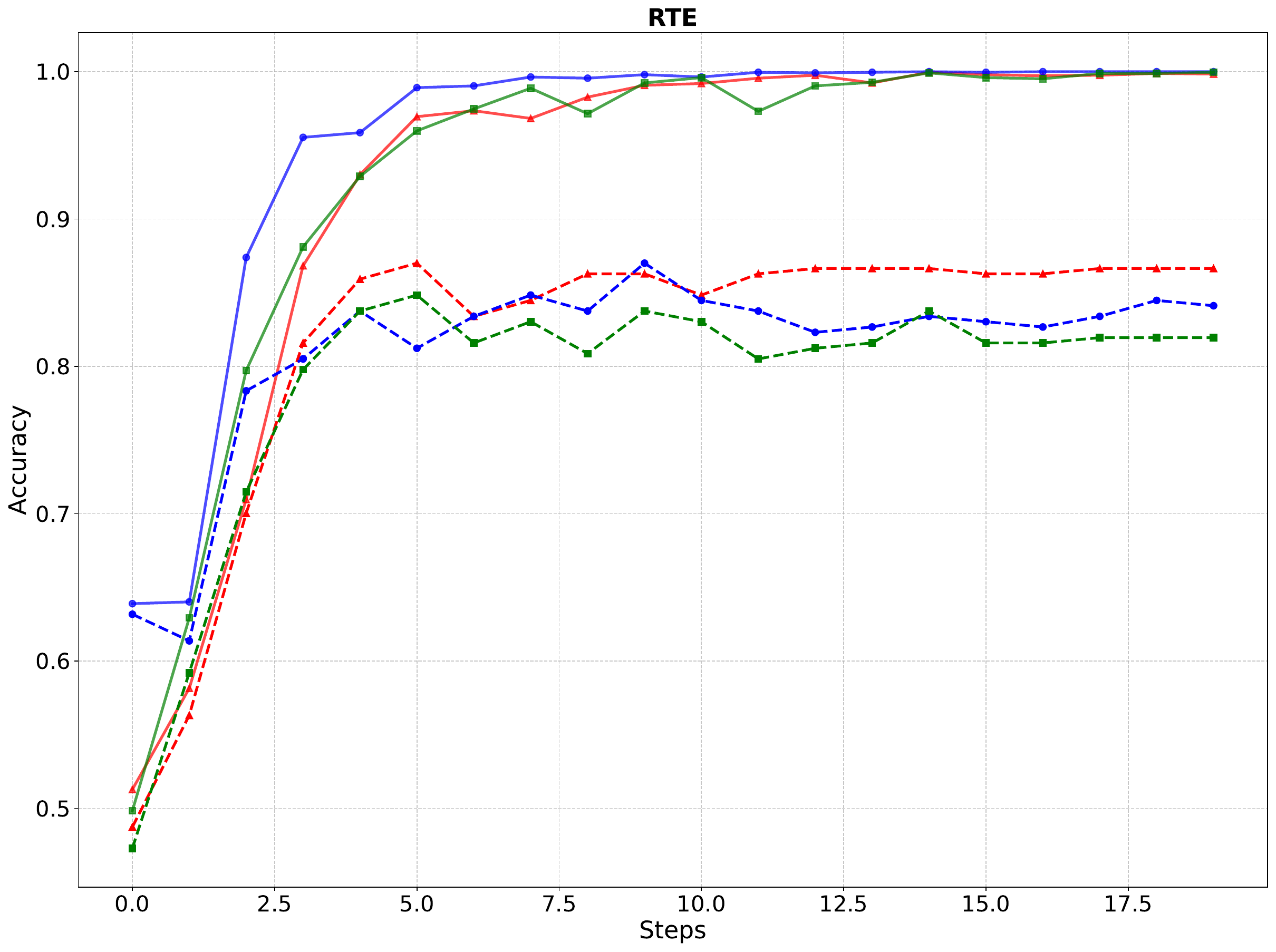}
        \caption{LoRMA$_\pi$ similar trend as others.}
    \end{subfigure}
    
    \begin{subfigure}[c]{\textwidth}
        \centering
        \includegraphics[width=0.6\textwidth]{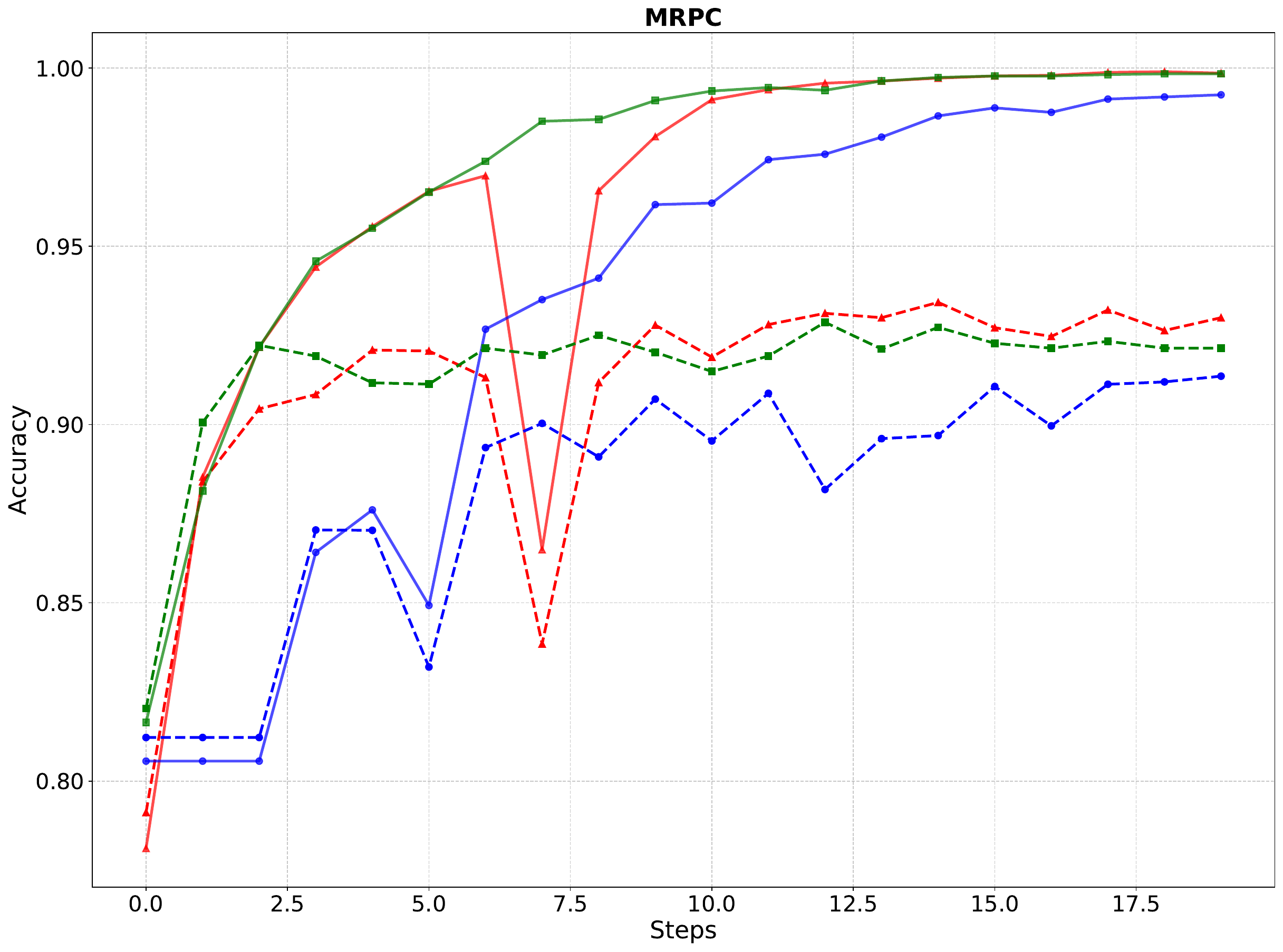}
        \caption{LoRMA$_\pi$ underfitting.}
    \end{subfigure}
    \caption{Training and Evaluation Accuracy variation for LoRA, LoRMA$_\pi$ and LoRMA$_+$.}   
    \label{fig:pi-vs-plus}
\end{figure*}

Overall, the performance of \modelshortShift, though comparable, is seen to be slightly lower than \modelshortRidge. We analyze this for various tasks, as shown in App. Fig. \ref{fig:pi-vs-plus}, and find the performance of \modelshortShift\ to require more task-specific tuning. For some tasks (a), \modelshortShift\ demonstrates a much better capacity to learn via train accuracy but requires task-specific regularization to translate this into test performance. For a few tasks (b), it performs similarly to \modelshortRidge. Finally, for a few runs (c), it seems susceptible to local minima. Therefore, deciding which of the two approaches to go with would depend on the task and hyper-parameter exploration budget.

\subsection{Rank Progression with training} \label{app:rank-progress}
\begin{figure}[h]
    \centering
    \includegraphics[width=\linewidth]{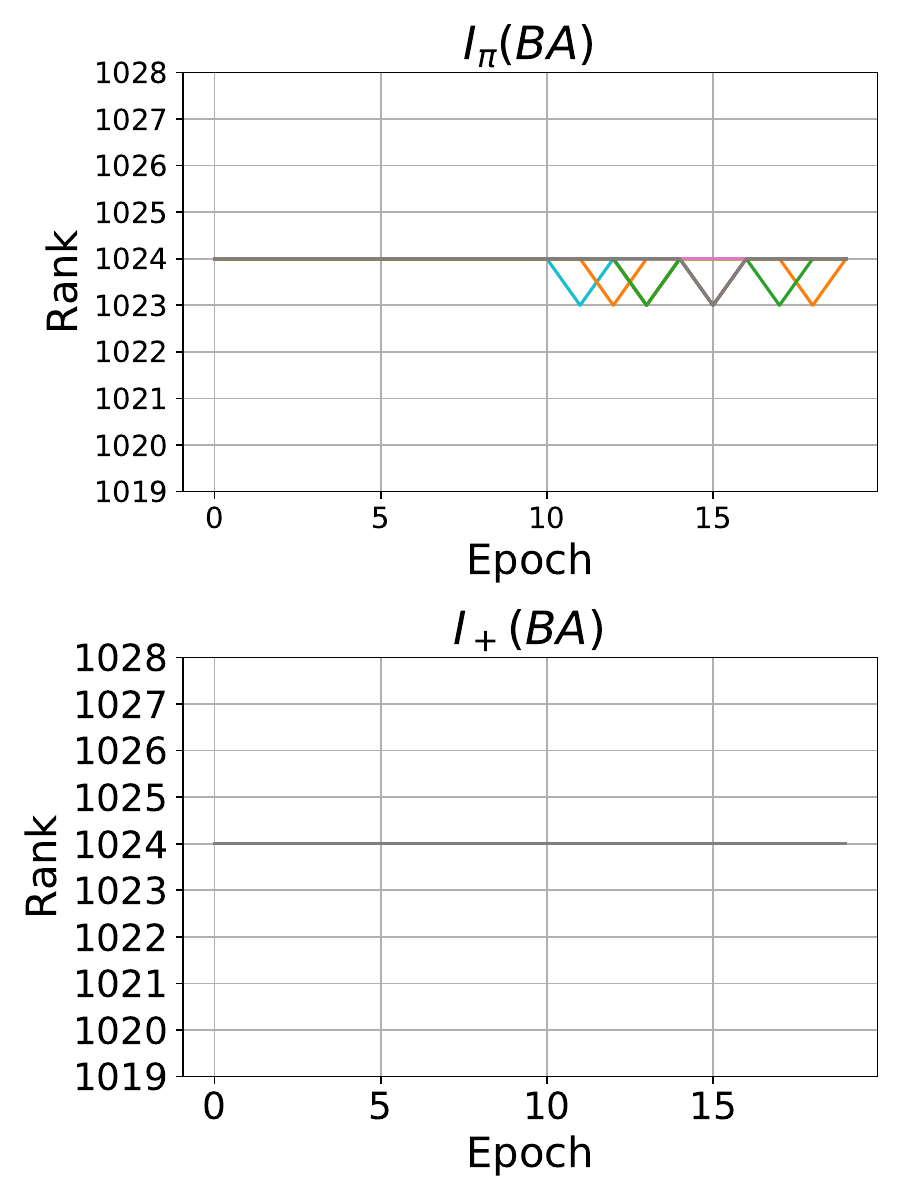}
    \caption{Variation of rank of resultant multiplicative adapters, i.e., $\mathcal{R}(\mathcal I_{\pi}(\mathbf B\mathbf A))$ and $\mathcal{R}(\mathcal I_{+}(\mathbf B\mathbf A))$  across epochs.}
    \label{fig:shift-rank-progress}
\end{figure}

As discussed in \S\ref{sec:methods}, the proposed initialization schemes, along with rank inflation, help to begin the training process with $\mathcal{I}(\mathbf{BA})=\mathbf{I}_d$ which is a full rank matrix. To empirically verify whether the rank inflation techniques, beginning with a high rank during initialization, also retain it across the training process, we monitor the rank $\mathcal{I}(\mathbf{BA})$ where $d=1024, \mathbf B \in \mathbb{R}^{1024 \times 8}, \mathbf A \in \mathbb{R} ^ {8 \times 1024} $. As depicted in Figure \ref{fig:shift-rank-progress} for both $\mathcal{I}_+$ and $\mathcal{I}_{\pi}$, throughout the fine-tuning process, it is observed that the inflated matrix product is always almost full rank (1024). Thus helping preserve the representational capacity of the matrix product.

\subsection{Ranks of the weight updates} \label{app:rank-of-weight-update}

To measure the richness of the weight updates received by the end of the fine-tuning process, we compare the ranks of $\Delta \mathbf W$. Table \ref{tab:weight-update-rank} shows the substantial difference in their ranks. In the case of LoRA, the weight update ($\Delta \mathbf W = \mathbf{BA}$) is constrained to be of rank $r=8$. A similar bound exists in the case of \modelshortRidge ($\Delta \mathbf W = \mathbf{BAW}_0$). For \modelshortShift, no such restriction exists, and the weight update is an almost full rank matrix.

\begin{table}
\centering
\small
\begin{tabular}{cc}
\toprule
Fine-Tuning Method & Rank of Weight Update \\
\midrule
LoRA & 8 \\
\modelshortRidge & 8 \\
\modelshortShift & 1021 \\
\bottomrule
\end{tabular}
\caption{Rank of the $\Delta \mathbf W$ for Layer 13 of RoBERTa\textsubscript{large} being fine-tuned for CoLA for $r=8$.}
\label{tab:weight-update-rank}
\end{table}

\begin{table}
\centering
\small
\begin{tabular}{l|ccc}
\hline
\toprule
Task & E2E & WebNLG & DART \\
\midrule
&\multicolumn{3}{c}{Training} \\
\midrule
Optimizer & \multicolumn{3}{c}{AdamW} \\
Weight Decay & 0.01 & 0.01 & 0.0\\
Dropout Prob & 0.1 & 0.1 & 0.0\\
Batch Size & \multicolumn{3}{c}{8} \\
\# Epoch & \multicolumn{3}{c}{5} \\
Warmup Steps & \multicolumn{3}{c}{500} \\
Learning Rate Schedule & \multicolumn{3}{c}{Linear} \\
Label Smooth & 0.1 & 0.1 & 0.0 \\
Learning Rate ($\mathcal I_+$) & \multicolumn{3}{c}{0.0002} \\
Learning Rate ($\mathcal I_\pi$) & \multicolumn{3}{c}{0.0001} \\
Matrices and $r$ & \multicolumn{3}{c}{$r_q=r_v=4$} \\
Scaling $\alpha$ ($\mathcal I_+$) & \multicolumn{3}{c}{32} \\
Scaling $\alpha$ ($\mathcal I_\pi$) & \multicolumn{3}{c}{8} \\
\midrule
&\multicolumn{3}{c}{Inference} \\
\midrule
Beam Size & 10/15 & 10 & 10 \\
Length Penalty & 0.9 & 0.8 & 0.8 \\
no repeat ngram size & \multicolumn{3}{c}{4} \\
\bottomrule
\end{tabular}
\caption{The hyperparameters for GPT-2 M LoRMA on E2E, WebNLG and DART.}
\label{tab:hyper_gpt2}
\end{table}

\begin{table}
    \centering
    \small
    \begin{tabular}{lcc}
        \toprule
         Hyperparams & Gemma-2B & LLaMA-3-8B \\
         \midrule
         Optimizer & \multicolumn{2}{c}{AdamW} \\
         Warmup Ratio & \multicolumn{2}{c}{0.1} \\
         LR Schedule & \multicolumn{2}{c}{Cosine} \\
         Learning Rate & 5E-4 & 5E4 \\
         Max Seq. Len. & \multicolumn{2}{c}{512} \\
        \# Epochs & \multicolumn{2}{c}{2} \\
        Batch Size & \multicolumn{2}{c}{64} \\
        Order & \multicolumn{2}{c}{pre-multiplication} \\
        Rank $(r)$ & 4 & 8 \\ 

        \bottomrule
         
    \end{tabular}

    \caption{Hyperparameters for \modelshortRidge\ training for MetaMath-40K fine tuning.}

    \label{tab:math_hyperparam}
\end{table}

% \AM{Font too small, can we make this figure similar style as Fig. \ref{fig:rank_comp}}}

\end{document}